\documentclass[10pt,twocolumn,letterpaper]{article}

\usepackage{cvpr}              

\usepackage{amsmath,amsfonts,bm}

\def\eqref#1{equation~\ref{#1}}

\def\1{\bm{1}}

\def\vb{{\bm{b}}}

\def\mM{{\bm{M}}}

\def\mW{{\bm{W}}}
\def\mX{{\bm{X}}}

\DeclareMathAlphabet{\mathsfit}{\encodingdefault}{\sfdefault}{m}{sl}
\SetMathAlphabet{\mathsfit}{bold}{\encodingdefault}{\sfdefault}{bx}{n}

\def\gA{{\mathcal{A}}}

\def\gD{{\mathcal{D}}}

\def\gL{{\mathcal{L}}}

\def\gN{{\mathcal{N}}}

\def\gQ{{\mathcal{Q}}}

\def\gX{{\mathcal{X}}}
\def\gY{{\mathcal{Y}}}
\def\gZ{{\mathcal{Z}}}

\def\sR{{\mathbb{R}}}

\usepackage[dvipsnames]{xcolor}

\definecolor{salmon}{RGB}{250, 128, 114}

\newlength\savewidth\newcommand\shline{\noalign{\global\savewidth\arrayrulewidth
  \global\arrayrulewidth 1pt}\hline\noalign{\global\arrayrulewidth\savewidth}}

\usepackage{pifont}
\newcommand{\cmark}{\ding{51}}%
\newcommand{\xmark}{\ding{55}}%

\definecolor{cvprblue}{rgb}{0.21,0.49,0.74}
\usepackage[pagebackref,breaklinks,colorlinks,citecolor=cvprblue]{hyperref}

\usepackage{xcolor}

\definecolor{citecolor}{HTML}{0071bc}
\definecolor{paleplum}{rgb}{0.8, 0.6, 0.8}
\hypersetup{
  colorlinks,
  citecolor=violet,
  linkcolor=red
}
\usepackage[accsupp]{axessibility}

\newcommand{\hy}[1]{\textcolor{black}{#1}}

 \makeatletter
\def\@fnsymbol#1{\ensuremath{\ifcase#1\or \dagger\or \ddagger\or
   \mathsection\or \mathparagraph\or \|\or **\or \dagger\dagger
   \or \ddagger\ddagger \else\@ctrerr\fi}}
\makeatother

\title{Efficient Stitchable Task Adaptation}

\author{Haoyu He  \quad Zizheng Pan \quad Jing Liu \quad Jianfei Cai \quad Bohan Zhuang\thanks{Corresponding author. E-mail: $\tt  bohan.zhuang@gmail.com$}\\[0.2cm]
ZIP Lab, Monash University, Australia
}

\begin{document}
\maketitle
\begin{abstract}
The paradigm of pre-training and fine-tuning has laid the foundation for deploying deep learning models. However, 
{most} fine-tuning methods are designed to meet a specific resource budget. Recently, {considering diverse deployment scenarios with various resource budgets}, SN-Net~\cite{pan2023stitchable} is introduced to quickly obtain numerous new networks (stitches) from the pre-trained models (anchors) in a model family via model stitching. Although promising, SN-Net confronts new challenges when adapting it to new target domains, including huge memory and storage requirements and a long and sub-optimal multistage adaptation process. In this work, we present a novel framework, \textbf{E}fficient \textbf{S}titchable \textbf{T}ask \textbf{A}daptation (ESTA), to efficiently produce a palette of fine-tuned models that adhere to diverse resource
constraints. Specifically, we first tailor parameter-efficient fine-tuning to share low-rank updates among the stitches while maintaining independent bias terms. In this way, we largely reduce fine-tuning memory burdens and mitigate the interference among stitches that arises in task adaptation. Furthermore, we streamline a simple yet effective one-stage deployment pipeline, which estimates the important stitches to deploy with training-time gradient statistics. By assigning higher sampling probabilities to important stitches, we also get a boosted Pareto frontier. Extensive experiments on 25 downstream visual recognition tasks demonstrate that our ESTA is capable of generating stitches with smooth accuracy-efficiency trade-offs and surpasses the direct SN-Net adaptation by remarkable margins with significantly lower training time and fewer trainable parameters. Furthermore, we demonstrate the flexibility and scalability of our ESTA framework by stitching LLMs from LLaMA family, obtaining chatbot stitches of assorted sizes\footnote{Source code is available at \url{https://github.com/ziplab/Stitched_LLaMA}.}.

\end{abstract}    
\section{Introduction}
\label{sec:intro}

The paradigm of pre-training and fine-tuning has underpinned modern applications in both vision and language. With off-the-shelf models pre-trained on large-scale datasets, the de-facto choice is vanilla full fine-tuning, which tunes all the model parameters with the downstream data.
To reduce memory footprint and avoid overfitting, an emerging trend is to tune a small proportion of the model parameters while freezing the majority ones with Parameter-Efficient Fine-Tuning (PEFT) ~\cite{jia2022vpt,houlsby2019parameter,hu2022lora}. However, both full fine-tuning and PEFT target an exclusive specific resource budget for each downstream task, while in reality, we often need to deploy multiple models simultaneously to meet various resource demands. This makes us ponder: \emph{What is an effective and efficient way to obtain a palette of fine-tuned models meeting different resource constraints?}

A natural approach is to compress a well-trained large model into numerous smaller ones~\cite{coates2011importance,hubara2017quantized,lecun1989optimal,han2015learning}. However, the computational cost grows linearly with the number of deployment scenarios. Following the once-for-all network~\cite{cai2019once,guo2020single}, a few works first pre-train a weight-sharing over-parameterized supernet, then adapt the supernet to the downstream tasks. Although promising, training supernets at scale with large datasets requires prohibitive computational resources, \eg, thousands of GPU training hours, which is practically infeasible.

\begin{table*}[t]
\centering
\renewcommand\arraystretch{1.2}
\caption{Comparison of frameworks for obtaining a palette of networks to meet different hardware efficiency constraints. Supernet-based methods NAT~\cite{lu2021neural} and TOFA~\cite{kundu2023tofa} necessitate a supernet training stage plus long deployment GPU hours and are restricted to small FLOPs ranges. SN-Net~\cite{pan2023stitchable} employs model stitching to achieve a wide FLOPs range with significantly reduced deployment GPU hours. Our ESTA framework further reduces the deployment GPU hours, fine-tuning memory, and trainable parameters, and is scalable to produce LLM stitches with billions of parameters (Section~\ref{subsec:llm}). The deployment GPU hours and fine-tuning memory for both SN-Net and ESTA are measured on a single NVIDIA GeForce RTX 3090 GPU when stitching ViT-Ti/S/B anchors~\cite{dosovitskiy2021an} with batch size 64. $^*$ indicates data is inferred from the training recipe of TOFA's supernet architecture FBNetV3~\cite{dai2021fbnetv3}.
}\vspace{-0.5em}
\label{tab:intro}
\resizebox{\linewidth}{!}{
\begin{tabular}{l|cccccccc}
Method & Adapting supernet & Deployment GPU hours & Fine-tuning memory & Trainable parameter $\#$ & FLOPs range (G) & LLM friendly  \\\shline
NAT~\cite{lu2021neural} & \xmark & $>$1,000 & - & $>$10M & $[0.2,0.6]$ & \xmark \\
TOFA~\cite{kundu2023tofa} & \cmark & $>$1,000$^*$ & - & $>$10M$^*$ & $[0.2,2.5]$ & \xmark \\
SN-Net~\cite{pan2023stitchable} & \xmark & 19.3 & 13,235M & 124.2M & $[1.3,17.6]$ & \xmark \\
ESTA (ours) & \xmark & 5.0 & 9,685M & 4.6M & $[1.3,17.6]$ & \cmark
\end{tabular}
}
\vspace{-1.5em}
\end{table*}

On the other hand, there are many open-source large models~\cite{touvron2023llama,he2022masked,bao2022beit} from communities such as HuggingFace~\cite{wolf2020transformers} that are pre-trained on large-scale datasets and ready to be downloaded. Inspired by the success of model stitching~\cite{lenc2015understanding,bansal2021revisiting,csiszarik2021similarity}, recently, Stitchable Neural Networks (SN-Net)~\cite{pan2023stitchable} has been proposed to stitch pre-trained models (anchors) of the same family to quickly obtain a set of candidate new networks (stitches) with different accuracy-efficiency trade-offs of a wide FLOPs range. However, the scope of SN-Net is limited to the pre-training classification task on the same source domain~\cite{russakovsky2015imagenet}. When directly adapting the standard approach of SN-Net to a target domain, it faces new challenges. Specifically, the straightforward way to employ SN-Net in task adaptation typically requires three stages: first adapting anchors individually to the target domain, followed by stitching the adapted anchors, and finally evaluating all stitches to find and deploy the ones on the Pareto Frontier. Such a three-stage adaptation process is expensive and sub-optimal. Moreover, adapting SN-Net typically needs to load and optimize all parameters of multiple models, leading to a daunting training-time memory cost. For instance, stitching the two smallest models in the LLaMA family~\cite{touvron2023llama,touvron2023llama} with SN-Net under standard training settings (32-bit floating point parameters and Adam optimizer on a single GPU) requires more than 80G Video RAM that cannot fit into high-end GPUs such as NVIDIA A100. SN-Net also requires saving a separate instance of the model family for each task, making the storage grow linearly with the number of deployed applications.

Therefore, in this paper, we introduce a novel Efficient Stitchable Task Adaptation (ESTA) framework. Specifically, built upon SN-Net~\cite{pan2023stitchable}, ESTA introduces two main designs to overcome the aforementioned issues. \emph{First}, to reduce the memory and storage costs, we tailor a Parameter-efficient Stitch fine-Tuning (PST) method, which incorporates the representative sparse fine-tuning technique  LoRA~\cite{hu2022lora} to keep the anchors and stitching layer weights frozen, while approximating their updates with trainable low-rank decomposition matrices. In addition to stitch-agnostic LoRA modules, we further introduce stitch-specific bias terms to alleviate the conflict among different stitches. \emph{Second}, we propose a simple one-stage deployment pipeline to simultaneously adapt and stitch the pre-trained anchors to the target domain. Moreover, we propose to estimate and accumulate an importance score for each stitch via the saliency pruning criterion~\cite{lee2018snip}. With the importance scores, we develop a novel stitch-sampling method to assign important stitches with higher sampling probabilities. More importantly, after fine-tuning, the stitch importance scores can be directly used to infer the best ones at the Pareto frontier for deployment. Table~\ref{tab:intro} compares different efficient frameworks for diverse deployments.

Overall, our paper has the following major contributions. 1) This is a pioneering work to investigate the problem of efficient fine-tuning of SN-Net for task adaptation. Our solution ESTA is a 
successful endeavor that delivers dozens of ready-to-deploy vision models up to 17.6G FLOPs in 5 GPU hours 
as well as multiple chatbot models with billions of parameters by stitching LLaMA models~\cite{touvron2023llama,touvron2023llama2} for the instruction-following task. 2) We devise a parameter-efficient stitch fine-tuning method that incorporates trainable stitch-agnostic LoRA modules and stitch-specific bias terms, which largely reduces the fine-tuning memory footprints while alleviating the interference issue among stitches. Moreover, we streamline a simple one-stage deployment pipeline with a novel task-specific stitch sampling strategy that greatly reduces the deployment time and improves Pareto frontiers. 3) Extensive experiments conducted on 25 downstream visual recognition tasks demonstrate that when stitching ViT-Ti/S/B~\cite{dosovitskiy2021an}, our ESTA achieves remarkable performance improvements compared to the direct SN-Net adaptation counterpart while utilizing significantly fewer trainable parameters.
\section{Related Work}
\noindent\textbf{Model stitching.} Model stitching~\cite{lenc2015understanding,bansal2021revisiting} targets connecting the bottom layers of one network to the top layers of another with a stitching layer. Following~\cite{lenc2015understanding}, model stitching is first employed to measure the similarities for the inner representations learned by deep neural networks~\cite{bansal2021revisiting,csiszarik2021similarity,hernandez2022model}. Previous works~\cite{bansal2021revisiting,csiszarik2021similarity} observe that trained networks with different initializations can be effectively stitched without incurring significant performance drop. Based on this observation, DeRy~\cite{yang2022deep} stitches the blocks dissected from different pre-trained models and assembles a new model based on their quantified similarity for
better performance. Herdt \etal~\cite{herdt2023model} propose interpreting the inner representations of a deep network by stitching it with a pre-trained GAN generator. Recently, from the perspective of flexible deployment, SN-Net~\cite{pan2023stitchable} stitches models of different sizes within a model family to cheaply produce numerous new networks with diverse accuracy and efficiency trade-offs in the pre-training task. In contrast to~\cite{pan2023stitchable}, our work specifically targets obtaining a palette of fine-tuned models for the task adaptation setting. To do so, we make key redesigns including a parameter-efficient stitch fine-tuning method and a simple one-stage deployment pipeline to overcome the efficiency challenges in SN-Net adaptation.

\noindent\textbf{Parameter-efficient fine-tuning.} Parameter-efficient fine-tuning~\cite{houlsby2019parameter,li2021prefix,hu2022lora,jia2022vpt} is a powerful alternative to vanilla full tine-tuning, which updates only a small number of parameters while freezing the majority ones. Freezing the majority of parameters allows us to optimize storage and reduces the burden on training GPU memory, as there is no need to store their gradients or other training time statistics. Recent research freezes the vast majority of parameters while fine-tuning 
either the parameters that are inherited in the backbone~\cite{yosinski2014transferable,zaken2022bitfit} or additionally added, in the form of learnable prompts~\cite{jia2022vpt,lester2021power}, low-rank bottleneck layers~\cite{houlsby2019parameter}, learnable scaling and shifting factors~\cite{lian2022scaling}, and separated small networks~\cite{sung2022lst}. Meanwhile, PEFT has been employed in numerous downstream applications. For instance, while freezing the most parameters, Polyhistor~\cite{liu2022polyhistor} learns a hyper-network to generate adapter weights~\cite{houlsby2019parameter} for multi-task adaptation, LLaMA-Adapter~\cite{zhang2023llama} inserts learnable prompts into Transformer layers for the instruction-following task, and ControlNet~\cite{zhang2023adding} employs the zero-initialized convolutions to fine-tune diffusion generative models~\cite{rombach2022high}. However, these works target only one exclusive resource budget and are not scalable to diverse deployment scenarios. Unlike other methods, our ESTA is crafted to yield multiple fine-tuned models with diverse capacities while jointly incorporating stitch-specific and stitch-agnostic lightweight trainable parameters.

\noindent\textbf{Fine-tuning with multiple models.} The presence of more large-scale models has unleashed the potential to utilize multiple models instead of a single one during fine-tuning. One line of work selects the best pre-trained model in the model zoo to fine-tune. To identify the best model, prior studies quantize the transferability of the pre-trained models ahead of fine-tuning by estimating their accuracy on the downstream tasks~\cite{eaton2008modeling,ammar2014automated,kornblith2019better} or the generalization capability to mitigate the domain gaps~\cite{nguyen2020leep, tran2019transferability,you2021logme,bolya2021scalable}. Another way is to do model selection using
an efficient online learning regime~\cite{xu2021nasoa}. Considering the diversity among the pre-trained models, research efforts have been devoted to effectively exploit their knowledge with feature aggregation~\cite{shu2021zoo,rusu2016progressive,liu2019knowledge}, model merging~\cite{wortsman2022model,izmailov2018averaging,dai2011greedy, matena2022merging}, or a mixture-of-experts architecture~\cite{shu2022hub}. In contrast to previous works that merely target better performance, our work employs off-the-shelf pre-trained model families to produce plentiful new models for diverse deployment 
requirements.

\section{Preliminaries}

\subsection{Model Stitching} 
Considering an $L$-layer feed-forward artificial neural network $f_{\theta_1}: \gX \rightarrow \gY$ parameterized by ${\theta_1}$, that maps any input from the input space $\gX$ to the output space $\gY$, $f_{\theta_1}$ can be denoted as a composition of functions that $f_{\theta_1}=f_L \circ \cdots \circ f_1$, where $\circ$ denotes the function composition. In model stitching, $f_{\theta_1}$ can be split up at the $l$-th layer into two portions of functions where $l\in [1, L-1]$. Given any input $\mX \subseteq \gX$, the first portion of the head functions maps $\mX$ to activations $\mX_l$ at layer $l$, \ie, $H_{{\theta_1},l}(\mX)=f_l \circ \cdots \circ f_1=\mX_l$. The second portion of the tail functions maps $\mX_l$ to the final output, \ie, $T_{{\theta_1},l}(\mX_l)=f_L \circ \cdots \circ f_{l+1}$. 

Given another pre-trained $M$-layer artificial neural network $f_{\theta_2}$ parameterized by ${\theta_2}$ that is split up at the $m$-th layer that $m\in [2, M]$, we can then employ a stitching layer $S: \gA_{{\theta_1}, l} \rightarrow \gA_{{\theta_2}, m}$ to map between the two activation spaces at layers $l$ and $m$. In this case, we obtain a new network parameterized by $\phi$ and the stitching layer parameters, \ie,
\begin{equation}
\vspace{-0.1em}
    \begin{aligned}
        F_{\phi, S}(\mX)=T_{{\theta_2}, m}\circ S \circ H_{{\theta_1}, l}(\mX),
    \end{aligned}
\label{eq:model_stitching}
\vspace{-1.5mm}
\end{equation}
where $\phi$ consists of partial parameters from both $\theta_1$ and $\theta_2$.

\begin{figure*}[t]
\begin{center}
\includegraphics[width=\linewidth]{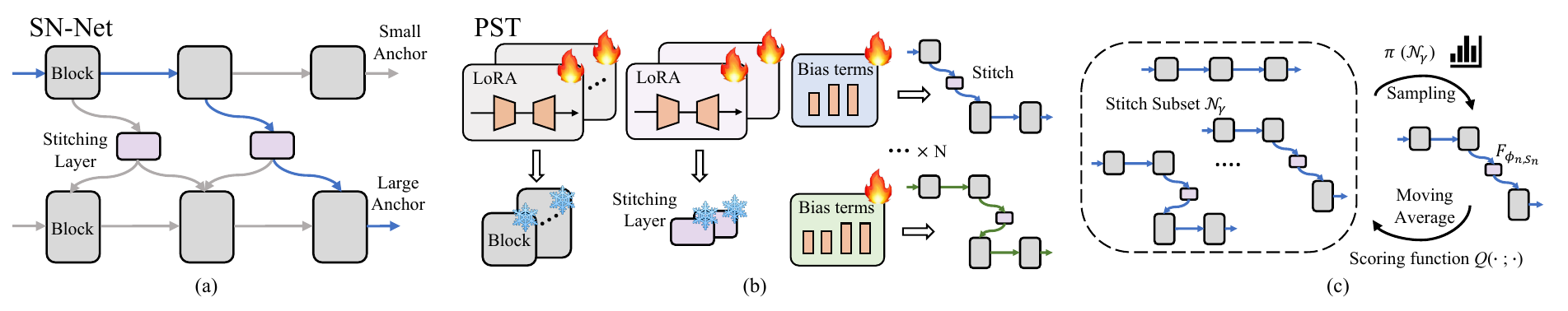}
\end{center}\vspace{-2em}
\caption{(a) Illustration of Stitchable Neural Network~\cite{pan2023stitchable}. With two anchors from the same model family, SN-Net connects the early layers of the smaller one to the latter layers of the larger one with stitching layers 
to obtain a set of new networks 
with different performance-efficiency trade-offs,
\eg, the path in Blue. (b) Overview of our PST method tailored for fine-tuning a palette of stitches, which integrates stitch-agnostic LoRA modules with stitch-specific bias terms, aiming to promote diverse representations among stitches while maintaining low trainable parameters. (c) Overview of our task-specific stitch sampling. We estimate the importance scores of the stitches with a scoring function $Q(\cdot, \cdot)$ and accumulate them as global statistics with moving averages. For a resource constraint $\gamma$, we sample with a categorical distribution $\pi(\gN_{\gamma})$ that is parameterized by the normalized importance scores so as to assign the important stitches with higher sampling probabilities. After fine-tuning, we directly deploy the stitches with the highest scores to avoid the costly evaluation stage.}\vspace{-1.5em}
\label{fig:main}
\end{figure*}

\begin{figure}[t]
\begin{center}
\includegraphics[width=\linewidth]{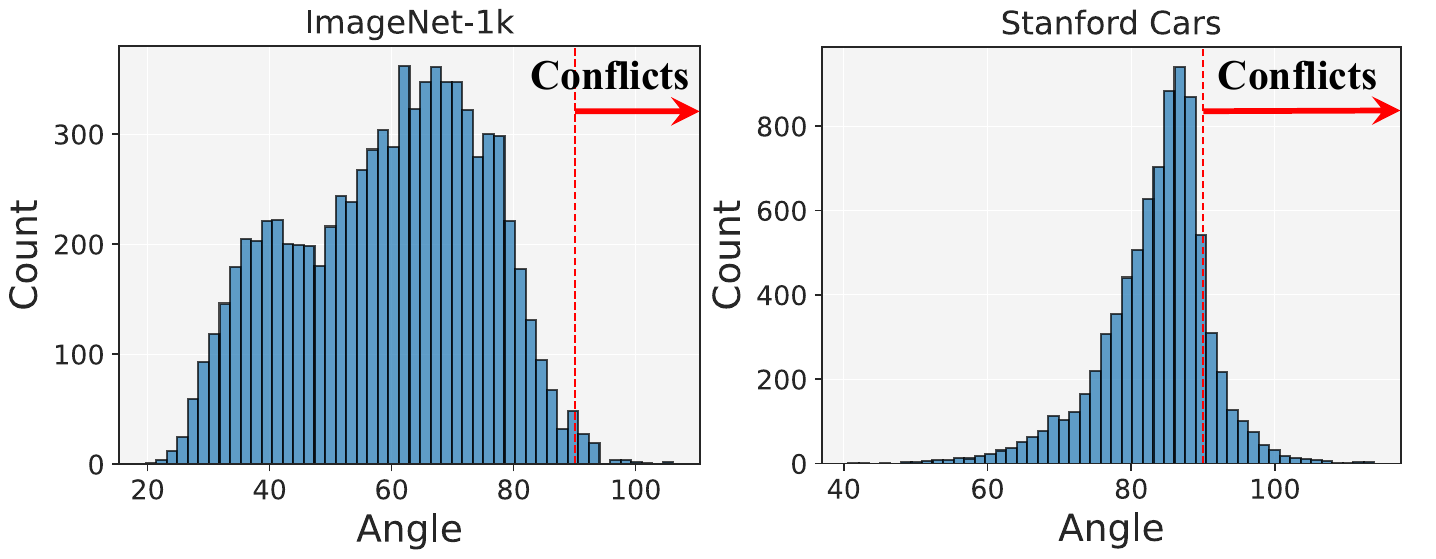}
\end{center}\vspace{-2em}
\caption{Distribution of pair-wise gradient angles among stitches when updating shared weights at fine-tuning iteration 600. We highlight angle 90$^\circ$ with a dashed red line. For simplicity, we show the gradient angles among the combined query, key, and value projection matrices for a total of 32 stitches when stitching ViT-Ti and ViT-S anchors. Generally, the gradient angles are larger in the target domain Stanford Cars~\cite{gebru2017cars} than in the source domain ImageNet-1k~\cite{russakovsky2015imagenet}.}
\label{fig:bias_motivation}
\vspace{-1.5em}
\end{figure}

\subsection{Stitchable Neural Network} 
In Stitchable Neural Network (SN-Net)~\cite{pan2023stitchable}, let a pre-trained model family of size $Z$ be $\gZ=\{f_{\theta_z}\}_{z=1}^Z$, where $\theta_z$ is the model parameter of the $z$-th model. The goal is to derive additional $N$ candidate new networks $\gN=\{F_{\phi_n, S_n}\}_{n=1}^N$ to adapt to various resource constraints. To do so, SN-Net first selects a pair of pre-trained models (anchors) from $\gZ$ and then stitches them with Eq.~(\ref{eq:model_stitching}) at different layer indexes to get $N$ stitches. Then, SN-Net samples the stitches randomly and jointly optimizes them. An overview of SN-Net is depicted in Figure~\ref{fig:main} (a). Since the anchors vary in scale, the newly assembled stitches have diverse performances and complexities. Importantly, SN-Net gives practical principles to design the space of $\gN$ that one should 1) stitch a pair of nearest anchors in terms of model complexity in a model family; 2) stitch the head of a faster and smaller anchor to the tail of a larger and slower anchor. Unless specified otherwise, we adopt these as the default experimental settings in this paper. 

\section{Method}

In this section, we introduce our ESTA framework, which consists of two major components:  \textit{parameter-efficient stitch fine-tuning} (PST) and \textit{a simple one-stage deployment pipeline}. PST is tailored to cheaply fine-tune a palette of stitches (Section~\ref{subsec:low_rank_update} and Figure~\ref{fig:main} (b)). The one-stage deployment pipeline aims to simultaneously save deployment time and improve adaptation performance (Section~\ref{subsec:one_stage}).

\subsection{Parameter-efficient Stitch Fine-tuning}\label{subsec:low_rank_update}

To address the problems of large storage and memory consumption when adapting SN-Net to downstream tasks with full fine-tuning, we introduce a parameter-efficient 
stitch fine-tuning (PST) method as depicted in Figure~\ref{fig:main} (b). Our basic idea is to adapt the representative sparse fine-tuning method LoRA~\cite{hu2022lora} for SN-Net fine-tuning.

\noindent\textbf{LoRA on Transformer layers}. Specifically, let any pre-trained weight matrix in the multi-head self-attention layer be $\mW\in \sR^{d\times k}$, we freeze $\mW$ and insert trainable low-rank decomposition matrices $\mW_{\rm down}\in \sR^{d\times r}$ and $\mW_{\rm up}\in\sR^{r\times k}$, where $r$ is the pre-defined rank that $r \ll {\rm min}(d, k)$. In this way, the updated version of $\mW$ can be formulated as
\begin{equation}
    \vspace{-0.1em}
    \begin{aligned}
        \mW'\leftarrow \mW + \mW_{\rm down}\mW_{\rm up}.
    \end{aligned}
\label{eq:update}
\end{equation}
We follow~\cite{hu2022lora} to respectively use Gaussian and zero initializations for $\mW_{\rm up}$ and $\mW_{\rm down}$, so that $\mW_{\rm down}\mW_{\rm up}$ is zero at the beginning of fine-tuning.

\noindent\textbf{LoRA on stitching layers}. We further extend the low-rank weight update to stitching layers. In SN-Net, a stitching layer is designed as a full-rank transformation matrix. Without loss of generality, we show the low-rank update of stitching layer $S_n$ that stitches $f_{\theta_1}$ and $f_{\theta_2}$ at layers $l$ and $m$, respectively. We first follow the initialization of stitching layers in~\cite{pan2023stitchable,csiszarik2021similarity} to align the activations at layers $l$ and $m-1$. Let the activations be $H_{\theta_1, l}(\mX)\in \sR^{b\times d_1}$ and $H_{{\theta_2}, m-1}(\mX)\in \sR^{b\times d_2}$ with sequence length $b$ and feature dimensions $d_1$ and $d_2$. The stitching layer is parameterized by a transformation matrix $\mM\in \mathbb{R}^{d_1\times d_2}$, which is initialized by solving a least squares problem, \ie, $\mM=H_{\theta_1, l}(\mX)^{\dagger}H_{{\theta_2}, m-1}(\mX)$, where $H_{\theta_1, l}(\mX)^{\dagger}$ is the Moore-Penrose pseudoinverse of $H_{\theta_1, l}(\mX)$. We find that such initialization already has impressive representational capacity on the target domain and they can be updated with low-rank decomposition without hurting the performance too much. Accordingly, we update $\mM$ similar to updating $\mW$ in Eq.~(\ref{eq:update}) by approximating its update with two learnable low-rank matrices with the same initializations.

Although the low-rank updates significantly enhance parameter efficiency, the low-rank essence~\cite{aghajanyan2020intrinsic} largely limits the network capacity. Particularly, in task adaptation, we often observe conflicting updates on the shared weights among the stitches. Figure~\ref{fig:bias_motivation} gives an example, where we stitch pre-trained anchors on ImageNet-1k~\cite{russakovsky2015imagenet} and Stanford Cars~\cite{gebru2017cars} and visualize the distribution of pair-wise gradient angles among different stitches when updating shared weights. We find that stitches agree less and conflict more (\ie, with more large angles) on the target domain Stanford Cars compared to the source domain ImageNet-1k.

\noindent\textbf{Stitch-specific bias}. 
To address the above issue, we further introduce stitch-specific bias terms as depicted in Figure~\ref{fig:main} (b). Particularly, for a linear layer parameterized by $\mW$ and a bias term $\vb$, we add bias term $\vb^s$ when optimizing the $s$-th stitch, for which the output activation becomes $\mW\mX + \vb + \vb^s$. We employ a set of distinct bias terms at different layers for each stitch. In this way, the stitches are encouraged to learn distinct feature representations for different resource requirements. To restrict the number of trainable parameters, following~\cite{hu2022lora}, we only apply PST to self-attention layers while freezing feed-forward layers.

\vspace{-0.5mm}
\subsection{One-stage Deployment Pipeline} \label{subsec:one_stage}
\noindent\textbf{Simultaneously adapt and stitch anchors.} As aforementioned, given a model family $\gZ$ pre-trained in the source domain, directly adapting SN-Net to a target domain typically involves three stages. First, adapting each anchor to the target domain $\gD$ by solving ${\theta_z}^*={\operatorname{argmin}}_{\theta_z}\gL(\theta_z; \gD$). In the second stage, at each training iteration, SN-Net samples a stitch and optimizes it with objective $\gL(\phi_n^*, S_n; \gD)$, where $\phi_n^*$ is the set of SN-Net parameters for stitch $F_{\phi_n, S_n}$ which has been optimized once in the first stage. In the final stage, all optimized stitches need to be evaluated to identify the ones on the Pareto frontier for diverse resource constraints. Such a three-stage approach is expensive and sub-optimal (see more discussions in the supplementary material). Thus, we propose to adapt and stitch the anchors simultaneously within one stage, by directly optimizing $\gL(\phi_n,S_n; \gD)$ for each sampled stitch. Moreover, we introduce a novel task-specific stitch-sampling method to allow the promising stitches more likely to be sampled, based on a stitch importance score. After fine-tuning, the stitches with the highest scores are naturally selected for deployment without the need for the above final stage evaluation.

\begin{figure*}[t]
\begin{center}
\includegraphics[width=\linewidth]{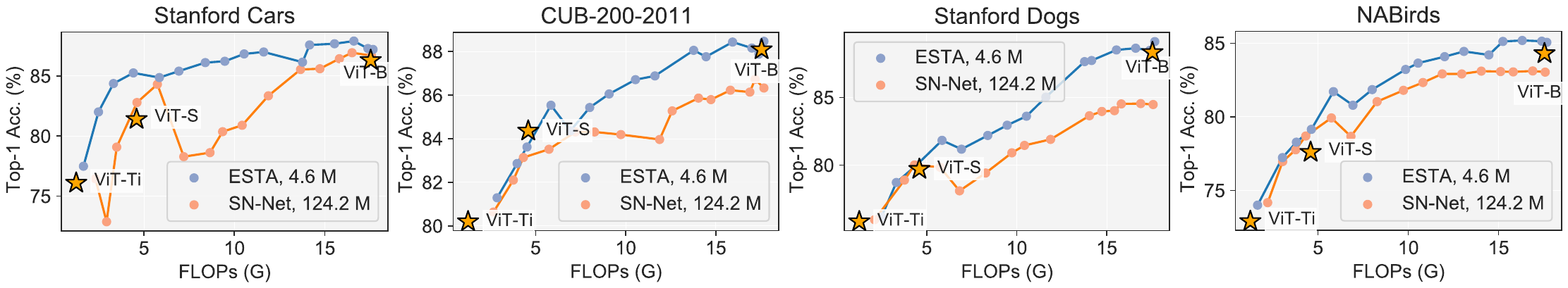}
\end{center}\vspace{-2em}
\caption{Performance comparisons with SN-Net~\cite{pan2023stitchable} for adapting ViT-Ti/S/B pre-trained on ImageNet-22k~\cite{deng2009imagenet} to Stanford Cars~\cite{gebru2017cars}, CUB-200-2011~\cite{wah2011caltech}, Stanford Dogs~\cite{Khosla_FGVC2011dogs}, and NABirds~\cite{van2015building}. We denote individually fine-tuned anchors as yellow stars. We also show the number of trainable parameters.}
\label{fig:main_results}
\vspace{-1.5em}
\end{figure*}

\begin{figure}[t]
\begin{center}
\includegraphics[width=\linewidth]{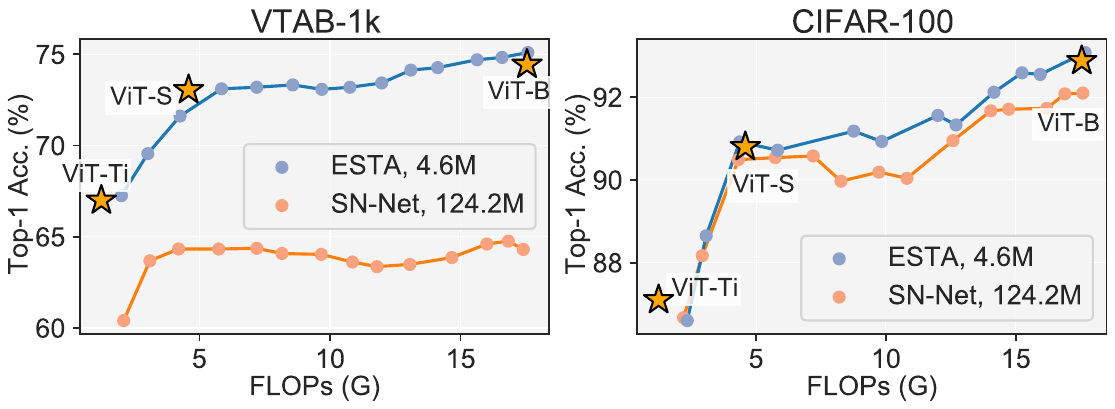}
\end{center}\vspace{-2.em}
\caption{Performance comparisons with SN-Net~\cite{pan2023stitchable} for adapting ViT-Ti/S/B pre-trained on ImageNet-22k~\cite{deng2009imagenet} to VTAB-1k~\cite{zhai2019vtab} and CIFAR-100~\cite{Krizhevsky09learningmultiple}. We denote individually fine-tuned anchors as yellow stars and also show the number of trainable parameters.}
\label{fig:main_vtab}
\vspace{-0.5em}
\end{figure}

\noindent\textbf{Task-specific stitch sampling.} SN-Net assumes that all stitches are equally important and performs random (uniform) sampling during training. However, as recognized by prior research~\cite{wang2021attentivenas,liu2022focusformer} that random sampling causes a gap between training and deployment as only the best stitches on the Pareto frontier are actually deployed. Moreover, since the importance of the pre-trained weights varies in different downstream tasks~\cite{guo2019spottune,xu2021raise,he2023sensitivity}, we argue that the performance of stitches that are parameterized by these weights is also different across these tasks.

To this end, we propose to assign the important stitches that are likely to be on the Pareto frontier with higher sampling probabilities during fine-tuning to ensure that they are optimized sufficiently. Specifically, we first estimate the importance of each stitch with a data-dependent saliency pruning metric SNIP~\cite{lee2018snip}, which measures the importance with the first-order gradient information with barely any extra computational cost. When the $n$-th stitch is sampled, we can get its importance score $Q(F_{\phi, S_n},\mX)$ with the scoring function $Q(\cdot, \cdot)$~\cite{lee2018snip}. To obtain robust importance scores for sampling, we accumulate scores with moving average during training, \ie,
\begin{equation}
\vspace{-0.1em}
    \begin{aligned}
        q_n^{t} \leftarrow \eta q_n^{t-1} + (1-\eta)Q(F_{\phi_n, S_n},\mX),
    \end{aligned}
\label{eq:moving_average}
\end{equation}
where $q_n^{t}$ and $q_n^{t-1}$ are the importance score at the $t$-th and $(t-1)$-th iteration, respectively, and $\eta\in [0,1)$ is the momentum coefficient. In this way, we can get stable importance scores. After a warm-up period that employs uniform sampling to accumulate the scores, we assign the important stitches with higher sampling probabilities for the rest of the fine-tuning epochs. To do so, we uniformly divide the resource constraint range of our stitches into several intervals and sample an interval $\gamma$ with uniform sampling. Accordingly, we can obtain a subset of stitches $\gN_\gamma$ whose resource constraints belong to $\gamma$ and their corresponding importance scores $\gQ_\gamma$. We can then define a categorical distribution based on the normalized importance scores, \ie, $\pi(\gN_\gamma)=\operatorname{softmax}(\gQ_\gamma)$ and finally get the sampled stitch $F_{\phi_n, S_n}\sim  \pi\left(\gN_\gamma\right)$. We show that our task-specific stitch sampling improves the performance in Section~\ref{subsec:abl}.

Additionally, with the sampled stitch, SN-Net employs knowledge distillation with a RegNetY-160~\cite{radosavovic2020designing} pre-trained teacher to improve its performance. However, pre-trained teachers on downstream tasks are often unavailable. Therefore, in each training iteration, we always sample and train the largest anchor (teacher) and transfer its knowledge to the sampled stitch (student) similar to inplace distillation~\cite{yu2019universally}. In practice, we employ the hard-label distillation~\cite{hinton2015distilling,touvron2021training}.

\noindent\textbf{Reuse stitch importance scores for deployment}. After fine-tuning, with the set of accumulated importance scores $\gQ_\gamma$ for all stitches, we directly use them to select the stitches with the highest scores to deploy, which eliminates the need for the costly evaluation of the final stage of the straightforward SN-Net adaptation solution. We empirically observe that the deployed stitches are mostly on the Pareto frontier in Section~\ref{subsec:abl}, suggesting our estimation of the important stitches is accurate. Furthermore, in the supplementary material, we show that the important stitches differ across different tasks, proving the need for task-specific deployment strategies, as highlighted in~\cite{guo2019spottune,xu2021raise,he2023sensitivity}. It's worth noting that although the SNIP metric has been widely employed as a one-shot metric to efficiently estimate network performance~\cite{abdelfattah2021zerocost,wang2020picking}, we extend it to estimate stitch importance for better training and selection.

\begin{figure*}[t]
\begin{center}
\includegraphics[width=\linewidth]{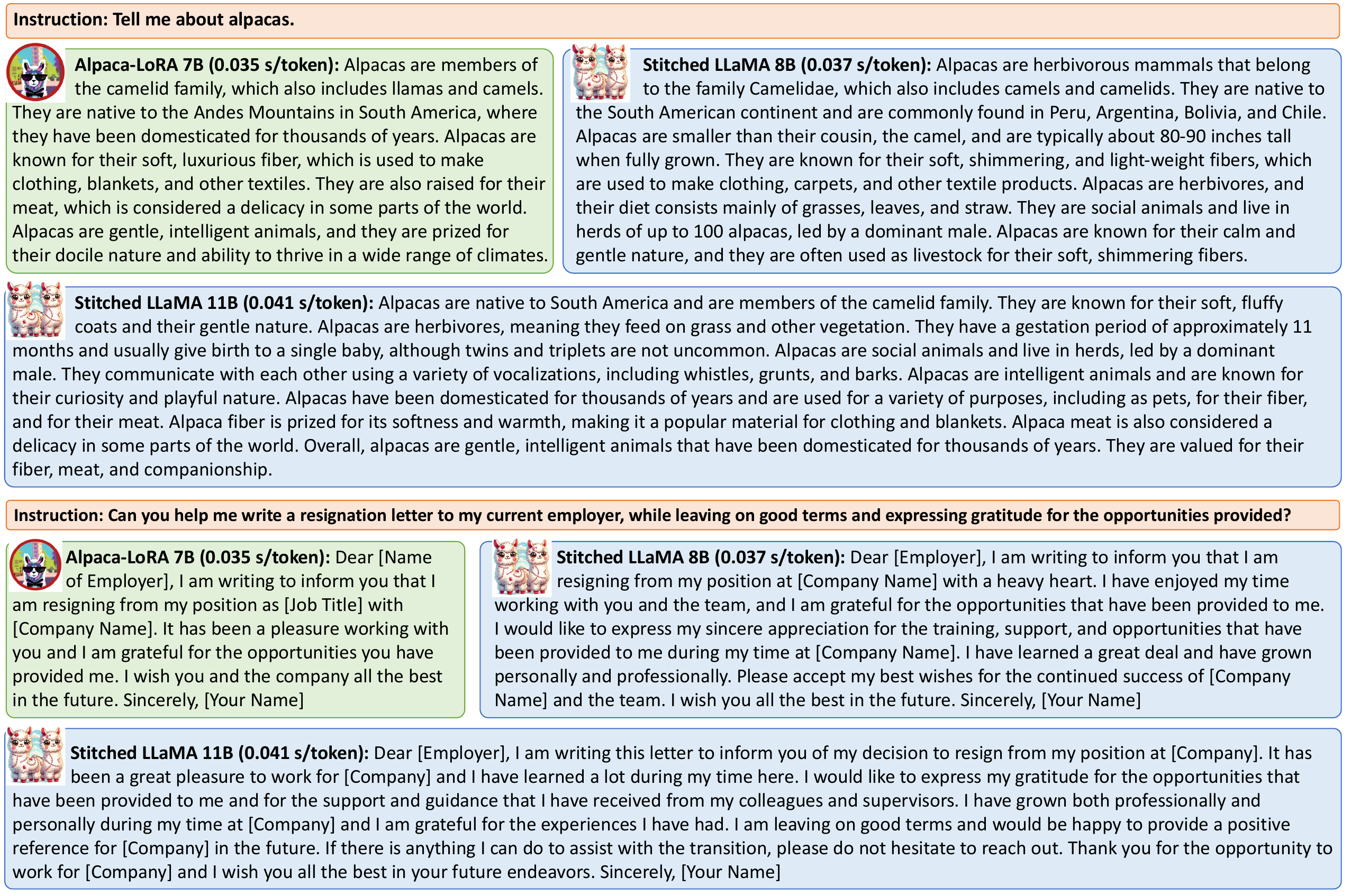}
\end{center}\vspace{-1.5em}
\caption{Instruction-following comparison between Stitched LLaMA obtained by our ESTA and the Alpaca-LoRA 7B fine-tuned with LoRA~\cite{hu2022lora}.}
\label{fig:llama}
\vspace{-0.5em}
\end{figure*}

\section{Experiments}

\subsection{Visual Recognition}
\label{subsec:visual_main}
We evaluate the effectiveness of our method on a total of 25 downstream visual recognition tasks, including fine-grained visual classification (FGVC) benchmark, common visual classification benchmark CIFAR-100~\cite{Krizhevsky09learningmultiple}, and VTAB-1k~\cite{zhai2019vtab} benchmark. FGVC benchmark contains NABirds~\cite{van2015building}, CUB-200-2011~\cite{wah2011caltech}, Stanford Cars~\cite{gebru2017cars}, Stanford Dogs~\cite{Khosla_FGVC2011dogs}, and Oxford Flowers~\cite{nilsback2008automated} tasks and the VTAB-1k benchmark includes 19 tasks in different domains, each of which has 800 training samples. For visual recognition experiments, we stitch ImageNet-21k pre-trained ViT-Ti/S/B models~\cite{dosovitskiy2021an} from~\cite{steiner2021train}. 
 
\noindent\textbf{Implementation details.} We employ the stitching space and settings (kernel size as $2$ and stride as $1$) in~\cite{pan2023stitchable} for downstream visual recognition tasks. We uniformly divide the resource constraints supported by the stitches into around 15 intervals and deploy one stitch within each interval. We set the hyper-parameter $r$ for the rank of weight updating in the self-attention layers to be $32$, $16$, and $8$ for ViT-Ti, ViT-S, and ViT-B, respectively. We also set $r=d_1//4$ universally for all stitching layers and $\eta$ in Eq.~(\ref{eq:moving_average}) as 0.9  by grid search. We follow~\cite{jia2022vpt} to fine-tune 100 epochs on each task and select the other hyper-parameters and augmentation methods. We include more implementation details in the supplementary material. 

\noindent\textbf{Main results.}
We compare the performance of our ESTA with SN-Net and the individual anchors fine-tuned with LoRA~\cite{hu2022lora} on various datasets. The FLOPs-accuracy curves are visualized in Figures~\ref{fig:main_results} and~\ref{fig:main_vtab}. We report the averaged results over the three task groups on VTAB-1k following~\cite{jia2022vpt}. Overall, the stitches obtained by our ESTA exhibit smooth FLOPs-accuracy trade-off curves consistently across all datasets. Notably, the curve for our ESTA outperforms SN-Net by significant margins with much fewer trainable parameters. We show that the inferior results of SN-Net on downstream tasks are led by the sub-optimal designs of fine-tuning all model parameters and a three-stage adaptation pipeline in Section~\ref{subsec:abl}.

We also observe that the stitches even outperform the anchors with comparable or lower FLOPs in many cases. For instance, stitches outperform individually fine-tuned ViT-S/B anchors by clear margins on NABirds, CUB-200-2011, Stanford Cars, and Stanford Dogs datasets with comparable computational complexity. This phenomenon is consistent with the observations made in~\cite{yu2019universally,pham2018efficient}. We conjecture that weight sharing among the stitches serves as a strong regularization to improve their generalization. We show more results on few-shot learning and stitching Convnets~\cite{liu2022convnet} in the supplementary material.

\begin{table}[t]
\centering
\renewcommand\arraystretch{1.2}
\caption{Relative response quality to Alpaca-LoRA 7B~\cite{alpaca} on Vicuna Bench~\cite{vicuna2023}. Our Stitched LLaMA successfully interpolates the answer quality between Alpaca-LoRA 7B and 13B.}
\vspace{-1em}
\label{tab:llama}
\resizebox{\linewidth}{!}{
\begin{tabular}{l|cc|ccccc} 
& \multicolumn{2}{|c|}{Alpaca-LoRA} & \multicolumn{5}{|c}{Stitched LLaMA} \\\shline
Parameter \# (B) & 6.7 & 13.0 & 6.9 & 7.8 & 11.5 & 12.5 & 13.0 \\
Quality (\%) & 100 & 123 & 113 & 114 & 115 & 120 & 123 \\
\end{tabular}
}\vspace{-1.5em}
\end{table}

\subsection{Instruction-following Task} 
\label{subsec:llm}
We also evaluate the instruction-following capability when stitching the LLaMA family~\cite{touvron2023llama} with our ESTA framework. We adapt and fine-tune LLaMA-7B/13B on the 52K instruction-following dataset from~\cite{alpaca} to obtain chatbot stitches of varying sizes. The 52K instruction-following data
is generated from 175 human-written instruction-output pairs with~\cite{wang2022self}. Each sample in the dataset is formulated by a task description, the context of the task, and the answer that is generated by GPT-3.5 (text-davinci-003)~\cite{brown2020language}.

\noindent\textbf{Implementation details.} We employ the stitching space and settings (kernel size as $1$ and stride as $1$) in~\cite{pan2023stitchable} for the instruction-following task and introduce 40 stitches. We uniformly divide the resource constraints supported by the stitches into 6 intervals and deploy one stitch within each interval. We set $\eta$ in Eq.~(\ref{eq:moving_average}) as 0.9. The rank $r$ is set to $16$ and $d_1//4$ for LoRA modules in self-attention and stitching layers, respectively.  We employ Adam optimizer and fine-tune with batch size 128 for 10 epochs with gradient accumulation. During the generation stage, we employ top-$p$ sampling to sample from the top subword candidates with the probability $p=0.75$, conduct the beam search, and set the beam size as $4$.

\noindent\textbf{Results.} We term the stitches obtained by our ESTA framework as Stitched LLaMA and compare them with Alpaca-LoRA 7B and 13B that respectively freeze the LLaMA 7B and 13B backbones and update the self-attention weights with LoRA modules~\cite{hu2022lora}. For quantitative results reported in Table~\ref{tab:llama}, we observe that similar to the visual recognition tasks, our Stitched LLaMA with different numbers of parameters generates answers with plausible quality in-between Alpaca-LoRA 7B and 13B. Figure~\ref{fig:llama} gives qualitative comparisons. Our Stitched LLaMAs produce reasonable responses, suggesting the potential of our ESTA for adapting large language models (LLMs) for more downstream language tasks. More examples can be found in the supplementary material.

\subsection{Ablation Study}
\vspace{-0.5mm}
\label{subsec:abl}

\begin{figure*}[t]
\begin{center}
    \includegraphics[width=\linewidth]{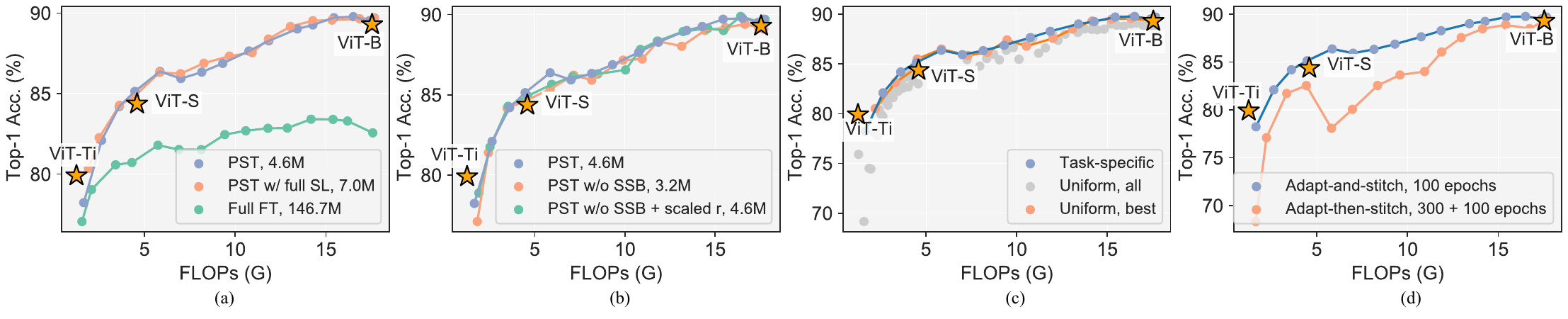}
\end{center}\vspace{-2.em}
\caption{Ablation studies with results averaged over five FGVC tasks. (a) Effect of our stitch-agnostic LoRA modules. ``Full FT'' represents full fine-tuning of all the model parameters.
``PST w/ full SL'' employs fully fine-tuned stitching layers. We also show the number of trainable parameters.  (b) Effect of our stitch-specific bias terms. ``SSB'' and ``scaled r'' represent stitch-specific bias terms and scaling the rank hyperparameter to reach 4.6M trainable parameters, respectively. (c) Effect of our task-specific stitch sampling. ``Uniform, all'' and ``Uniform, best'' represent all the stitches fine-tuned with uniform sampling and their best ones in each resource constraint interval on the Pareto frontier, respectively. (d) Effect of our strategy to simultaneously adapt and stitch anchors. Our strategy ``Adapt-and-stitch'' takes 100 epochs for fine-tuning. In contrast, ``Adapt-then-stitch'', as a straightforward approach to apply SN-Net, first individually adapts each anchor for a total of $300$ epochs, and then fine-tunes SN-Net for another 100 epochs.}
\label{fig:abl_effect}\vspace{-1.5em}
\end{figure*}

\noindent\textbf{Effect of parameter-efficient stitch fine-tuning.}
We investigate the effectiveness of our PST on five FGVC tasks and visualize the averaged results in Figure~\ref{fig:abl_effect} (a) and (b). 
To recall, our PST keeps the majority of the model parameters frozen while introducing trainable stitch-agonistic low-rank decomposition matrices and stitch-specific bias terms. In Figure~\ref{fig:abl_effect} (a), we observe that when employing LoRA to the multi-head self-attention layers, there are solid performance gains on all FLOPs ranges with significantly fewer trainable parameters, echoing the observations in~\cite{jia2022vpt,lian2022scaling}. It is suggested that low-rank weight updates is a powerful alternative to full fine-tuning for stitching and adapting anchors and we speculate that low-rank weight update avoids overfitting under limited downstream data. We further employ low-rank weight updates to the stitching layers and save 2.4M more trainable parameters with a slight overall performance drop. We conjecture that the stitching layers for large pre-trained models are also in low intrinsic dimensions~\cite{aghajanyan2020intrinsic} and least-squares initialization~\cite{pan2023stitchable} already provide a good initialization for them.

Figure~\ref{fig:abl_effect} (b) shows that employing stitch-specific bias terms achieves better performance with affordable 1.4M extra trainable parameters. 
We also include the baseline that simply scales the low-rank dimension $r$ to reach the same number of trainable parameters as our PST. This baseline is inferior to our PST.
We speculate that the stitch-specific bias terms enable flexible adjustments of feature representations for different stitches, leading to enhanced performance.

\noindent\textbf{Effect of task-specific stitch sampling.} We investigate the effectiveness of our sampling strategy (introduced in Section~\ref{subsec:one_stage}) on the five FGVC tasks. The averaged results are visualized in Figure~\ref{fig:abl_effect} (c). We compare our task-specific stitch sampling with the uniform one that is adopted in~\cite{pan2023stitchable}. We empirically find that for uniform sampling, there is a clear performance gap between the best stitches that will be deployed and the others that will be dropped after training, suggesting the need for concentrating on the important stitches during training. To this end, our task-specific sampling strategy performs better than uniform sampling, indicating that our sampling strategy selects accurate task-specific important stitches. Most importantly, our strategy circumvents the need for the costly evaluation stage in the direct solution of adapting SN-Net for downstream tasks.

\noindent\textbf{Effect of simultaneously adapting and stitching anchors.}
We study the effect of simultaneously adapting and stitching anchors strategy (introduced in Section~\ref{subsec:one_stage}) on five FGVC tasks. The averaged results are visualized in Figure~\ref{fig:abl_effect} (d). We compare our approach with the direct SN-Net adaptation that first adapts each anchor individually for a total of 300 epochs before stitching them under the exact setting as ESTA. It can be seen that the direct SN-Net adaptation yields poor results in task adaptation. Fine-tuned individual anchors, especially the larger ones, are likely to overfit the limited downstream data~\cite{plested2022deep,liu2023pretrain}. Accordingly, we speculate that the adapted anchor weights serve as a sub-optimal initialization for the stitches in the context of model stitching. On the other hand, our approach achieves better performance and greatly saves the fine-tuning cost.

\begin{figure}[t]
\begin{center}
\includegraphics[width=\linewidth]{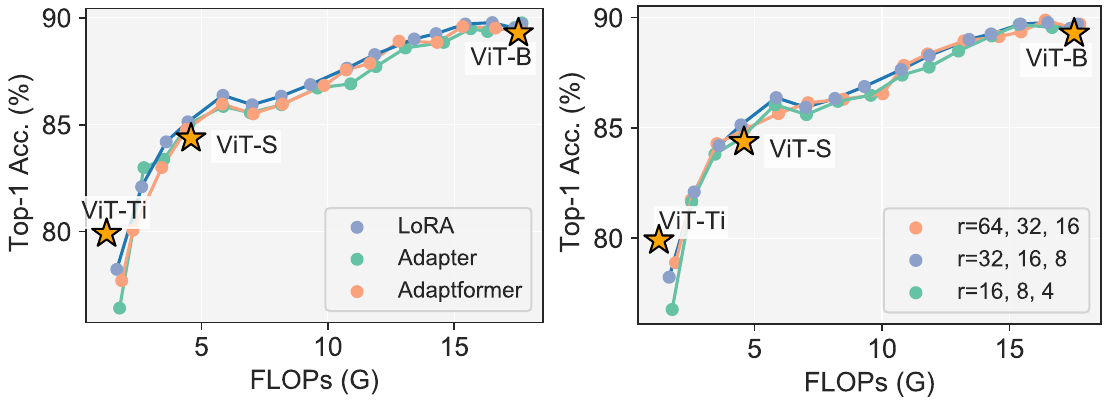}
\end{center}\vspace{-1.em}
\caption{Left: effect of different PEFT technique choices. Right: effect of the rank hyper-parameter $r$ in LoRA modules of self-attention layers, for which the three values represent the ranks in ViT-Ti, ViT-S, and ViT-B, respectively. The results are averaged on five FGVC tasks.}\vspace{-1.5em}
\label{fig:abl_other_peft}
\end{figure}

\noindent\textbf{Effect of the PEFT technique choices.} We compare the effect of different PEFT technique choices and visualize the averaged results on the five FGVC tasks in Figure~\ref{fig:abl_other_peft} left. We experiment to employ other PEFT techniques Adapter~\cite{houlsby2019parameter} and Adaptformer~\cite{chen2022adaptformer}. We follow their default setting to respectively insert bottleneck structures sequentially and in parallel to the feed-forward layers and fix the number of trainable parameters the same as ours for a fair comparison. We observe that the overall performance for different PEFT techniques is similar, suggesting that our ESTA framework is compatible with more PEFT technique choices. Since low-rank weight updates of LoRA~\cite{hu2022lora} can be merged into the backbone after training without extra inference computational cost, we employ LoRA by default.

\noindent\textbf{Effect of the low-rank hyper-parameter $r$.} 
We investigate the effect of different $r$ which controls the rank when updating weights in self-attention layers on the five FGVC tasks. The averaged results on the five FGVC tasks are visualized in Figure~\ref{fig:abl_other_peft} right. We find that setting $r$ to be different values barely has any effect on the overall performance, suggesting our ESTA framework is robust to the rank hyper-parameter $r$. Since setting $r$ to be $32$, $16$, and $8$ for low-rank updating the self-attention weights in ViT-Ti, ViT-S, and ViT-B has slightly higher averaged performance among the stitches, we make it our default setting.

\section{Conclusion and Future Work}

In this paper, we have introduced a novel task adaptation framework to cheaply obtain a palette of fine-tuned networks via model stitching, supporting diverse efficiency-performance tradeoffs at runtime.
Specifically, built on SN-Net, we have tailored a parameter-efficient stitch fine-tuning method, which learns lightweight stitch-agnostic LoRA modules and stitch-specific bias terms while keeping the majority of the parameters frozen. Our design significantly reduces fine-tuning memory and storage costs for downstream task adaptation. Moreover, we have devised a task-specific stitch sampling strategy to assign higher sampling probability to the important stitches during training, which simultaneously improves the Pareto frontiers and avoids a costly evaluation stage. Extensive experiments on 25 downstream visual recognition tasks and the instruction-following task have demonstrated the effectiveness of our proposed framework.

\noindent\textbf{Limitations.} Due to the constraints of computational resources, our experiments are limited to visual recognition and instruction-tuning tasks. In the future, we will explore adapting pre-trained model families to dense prediction and multi-modal tasks.

{
    \bibliographystyle{ieeenat_fullname}
    \small
    \bibliography{main}

\begin{thebibliography}{89}
\providecommand{\natexlab}[1]{#1}
\providecommand{\url}[1]{\texttt{#1}}
\expandafter\ifx\csname urlstyle\endcsname\relax
  \providecommand{\doi}[1]{doi: #1}\else
  \providecommand{\doi}{doi: \begingroup \urlstyle{rm}\Url}\fi

\bibitem[Abdelfattah et~al.(2021)Abdelfattah, Mehrotra, Dudziak, and Lane]{abdelfattah2021zerocost}
Mohamed~S. Abdelfattah, Abhinav Mehrotra, Lukasz Dudziak, and Nicholas~D. Lane.
\newblock Zero-cost proxies for lightweight nas.
\newblock In \emph{ICLR}, 2021.

\bibitem[Aghajanyan et~al.(2021)Aghajanyan, Zettlemoyer, and Gupta]{aghajanyan2020intrinsic}
Armen Aghajanyan, Luke Zettlemoyer, and Sonal Gupta.
\newblock Intrinsic dimensionality explains the effectiveness of language model fine-tuning.
\newblock In \emph{ACL}, 2021.

\bibitem[Ammar et~al.(2014)Ammar, Eaton, Taylor, Mocanu, Driessens, Weiss, and Tuyls]{ammar2014automated}
Haitham~Bou Ammar, Eric Eaton, Matthew~E Taylor, Decebal~Constantin Mocanu, Kurt Driessens, Gerhard Weiss, and Karl Tuyls.
\newblock An automated measure of mdp similarity for transfer in reinforcement learning.
\newblock In \emph{AAAIW}, 2014.

\bibitem[Andreassen et~al.(2021)Andreassen, Bahri, Neyshabur, and Roelofs]{andreassen2021evolution}
Anders Andreassen, Yasaman Bahri, Behnam Neyshabur, and Rebecca Roelofs.
\newblock The evolution of out-of-distribution robustness throughout fine-tuning.
\newblock \emph{arXiv preprint arXiv:2106.15831}, 2021.

\bibitem[Bansal et~al.(2021)Bansal, Nakkiran, and Barak]{bansal2021revisiting}
Yamini Bansal, Preetum Nakkiran, and Boaz Barak.
\newblock Revisiting model stitching to compare neural representations.
\newblock \emph{NeurIPS}, 34:\penalty0 225--236, 2021.

\bibitem[Bao et~al.(2022)Bao, Dong, Piao, and Wei]{bao2022beit}
Hangbo Bao, Li Dong, Songhao Piao, and Furu Wei.
\newblock Beit: Bert pre-training of image transformers.
\newblock In \emph{ICLR}, 2022.

\bibitem[Bolya et~al.(2021)Bolya, Mittapalli, and Hoffman]{bolya2021scalable}
Daniel Bolya, Rohit Mittapalli, and Judy Hoffman.
\newblock Scalable diverse model selection for accessible transfer learning.
\newblock \emph{NeurIPS}, 34:\penalty0 19301--19312, 2021.

\bibitem[Brown et~al.(2020)Brown, Mann, Ryder, Subbiah, Kaplan, Dhariwal, Neelakantan, Shyam, Sastry, Askell, et~al.]{brown2020language}
Tom Brown, Benjamin Mann, Nick Ryder, Melanie Subbiah, Jared~D Kaplan, Prafulla Dhariwal, Arvind Neelakantan, Pranav Shyam, Girish Sastry, Amanda Askell, et~al.
\newblock Language models are few-shot learners.
\newblock \emph{NeurIPS}, 33:\penalty0 1877--1901, 2020.

\bibitem[Cai et~al.(2020)Cai, Gan, Wang, Zhang, and Han]{cai2019once}
Han Cai, Chuang Gan, Tianzhe Wang, Zhekai Zhang, and Song Han.
\newblock Once-for-all: Train one network and specialize it for efficient deployment.
\newblock In \emph{ICLR}, 2020.

\bibitem[Chen et~al.(2022)Chen, Ge, Tong, Wang, Song, Wang, and Luo]{chen2022adaptformer}
Shoufa Chen, Chongjian Ge, Zhan Tong, Jiangliu Wang, Yibing Song, Jue Wang, and Ping Luo.
\newblock Adaptformer: Adapting vision transformers for scalable visual recognition.
\newblock \emph{NeurIPS}, 2022.

\bibitem[Chiang et~al.(2023)Chiang, Li, Lin, Sheng, Wu, Zhang, Zheng, Zhuang, Zhuang, Gonzalez, Stoica, and Xing]{vicuna2023}
Wei-Lin Chiang, Zhuohan Li, Zi Lin, Ying Sheng, Zhanghao Wu, Hao Zhang, Lianmin Zheng, Siyuan Zhuang, Yonghao Zhuang, Joseph~E. Gonzalez, Ion Stoica, and Eric~P. Xing.
\newblock Vicuna: An open-source chatbot impressing gpt-4 with 90\%* chatgpt quality, 2023.

\bibitem[Coates and Ng(2011)]{coates2011importance}
Adam Coates and Andrew~Y Ng.
\newblock The importance of encoding versus training with sparse coding and vector quantization.
\newblock In \emph{ICML}, pages 921--928, 2011.

\bibitem[Csisz{\'a}rik et~al.(2021)Csisz{\'a}rik, K{\H{o}}r{\"o}si-Szab{\'o}, Matszangosz, Papp, and Varga]{csiszarik2021similarity}
Adri{\'a}n Csisz{\'a}rik, P{\'e}ter K{\H{o}}r{\"o}si-Szab{\'o}, Akos Matszangosz, Gergely Papp, and D{\'a}niel Varga.
\newblock Similarity and matching of neural network representations.
\newblock \emph{NeurIPS}, 34:\penalty0 5656--5668, 2021.

\bibitem[Dai and Zhang(2011)]{dai2011greedy}
Dong Dai and Tong Zhang.
\newblock Greedy model averaging.
\newblock \emph{NeurIPS}, 24, 2011.

\bibitem[Dai et~al.(2021)Dai, Wan, Zhang, Wu, He, Wei, Chen, Tian, Yu, Vajda, et~al.]{dai2021fbnetv3}
Xiaoliang Dai, Alvin Wan, Peizhao Zhang, Bichen Wu, Zijian He, Zhen Wei, Kan Chen, Yuandong Tian, Matthew Yu, Peter Vajda, et~al.
\newblock Fbnetv3: Joint architecture-recipe search using predictor pretraining.
\newblock In \emph{CVPR}, pages 16276--16285, 2021.

\bibitem[Deng et~al.(2009)Deng, Dong, Socher, Li, Li, and Fei-Fei]{deng2009imagenet}
Jia Deng, Wei Dong, Richard Socher, Li-Jia Li, Kai Li, and Li Fei-Fei.
\newblock Imagenet: A large-scale hierarchical image database.
\newblock In \emph{CVPR}, pages 248--255, 2009.

\bibitem[Dettmers et~al.(2022)Dettmers, Lewis, Belkada, and Zettlemoyer]{dettmers2022llm}
Tim Dettmers, Mike Lewis, Younes Belkada, and Luke Zettlemoyer.
\newblock Llm. int8 (): 8-bit matrix multiplication for transformers at scale.
\newblock \emph{NeurIPS}, 2022.

\bibitem[Dosovitskiy et~al.(2021)Dosovitskiy, Beyer, Kolesnikov, Weissenborn, Zhai, Unterthiner, Dehghani, Minderer, Heigold, Gelly, Uszkoreit, and Houlsby]{dosovitskiy2021an}
Alexey Dosovitskiy, Lucas Beyer, Alexander Kolesnikov, Dirk Weissenborn, Xiaohua Zhai, Thomas Unterthiner, Mostafa Dehghani, Matthias Minderer, Georg Heigold, Sylvain Gelly, Jakob Uszkoreit, and Neil Houlsby.
\newblock An image is worth 16x16 words: Transformers for image recognition at scale.
\newblock In \emph{ICLR}, 2021.

\bibitem[Eaton et~al.(2008)Eaton, Desjardins, and Lane]{eaton2008modeling}
Eric Eaton, Marie Desjardins, and Terran Lane.
\newblock Modeling transfer relationships between learning tasks for improved inductive transfer.
\newblock In \emph{ECML PKDD}, pages 317--332. Springer, 2008.

\bibitem[Gebru et~al.(2017)Gebru, Krause, Wang, Chen, Deng, and Fei-Fei]{gebru2017cars}
Timnit Gebru, Jonathan Krause, Yilun Wang, Duyun Chen, Jia Deng, and Li Fei-Fei.
\newblock Fine-grained car detection for visual census estimation.
\newblock In \emph{AAAI}, 2017.

\bibitem[Guo et~al.(2019)Guo, Shi, Kumar, Grauman, Rosing, and Feris]{guo2019spottune}
Yunhui Guo, Honghui Shi, Abhishek Kumar, Kristen Grauman, Tajana Rosing, and Rogerio Feris.
\newblock Spottune: transfer learning through adaptive fine-tuning.
\newblock In \emph{CVPR}, pages 4805--4814, 2019.

\bibitem[Guo et~al.(2020)Guo, Zhang, Mu, Heng, Liu, Wei, and Sun]{guo2020single}
Zichao Guo, Xiangyu Zhang, Haoyuan Mu, Wen Heng, Zechun Liu, Yichen Wei, and Jian Sun.
\newblock Single path one-shot neural architecture search with uniform sampling.
\newblock In \emph{ECCV}, pages 544--560. Springer, 2020.

\bibitem[Han et~al.(2015)Han, Pool, Tran, and Dally]{han2015learning}
Song Han, Jeff Pool, John Tran, and William Dally.
\newblock Learning both weights and connections for efficient neural network.
\newblock \emph{NeurIPS}, 28, 2015.

\bibitem[He et~al.(2023)He, Cai, Zhang, Tao, and Zhuang]{he2023sensitivity}
Haoyu He, Jianfei Cai, Jing Zhang, Dacheng Tao, and Bohan Zhuang.
\newblock Sensitivity-aware visual parameter-efficient tuning.
\newblock In \emph{ICCV}, 2023.

\bibitem[He et~al.(2022)He, Chen, Xie, Li, Doll{\'a}r, and Girshick]{he2022masked}
Kaiming He, Xinlei Chen, Saining Xie, Yanghao Li, Piotr Doll{\'a}r, and Ross Girshick.
\newblock Masked autoencoders are scalable vision learners.
\newblock In \emph{CVPR}, pages 16000--16009, 2022.

\bibitem[Herdt et~al.(2023)Herdt, Schmidt, Baguer, Arrastia, and Maass]{herdt2023model}
Rudolf Herdt, Maximilian Schmidt, Daniel~Otero Baguer, Jean~Le'Clerc Arrastia, and Peter Maass.
\newblock Model stitching and visualization how gan generators can invert networks in real-time.
\newblock \emph{arXiv preprint arXiv:2302.02181}, 2023.

\bibitem[Hernandez et~al.(2022)Hernandez, Dangovski, Lu, and Soljacic]{hernandez2022model}
Adriano Hernandez, Rumen Dangovski, Peter~Y. Lu, and Marin Soljacic.
\newblock Model stitching: Looking for functional similarity between representations.
\newblock \emph{NeurIPSW}, 2022.

\bibitem[Hinton et~al.(2014)Hinton, Vinyals, and Dean]{hinton2015distilling}
Geoffrey Hinton, Oriol Vinyals, and Jeff Dean.
\newblock Distilling the knowledge in a neural network.
\newblock \emph{NeurIPSW}, 2014.

\bibitem[Houlsby et~al.(2019)Houlsby, Giurgiu, Jastrzebski, Morrone, De~Laroussilhe, Gesmundo, Attariyan, and Gelly]{houlsby2019parameter}
Neil Houlsby, Andrei Giurgiu, Stanislaw Jastrzebski, Bruna Morrone, Quentin De~Laroussilhe, Andrea Gesmundo, Mona Attariyan, and Sylvain Gelly.
\newblock Parameter-efficient transfer learning for nlp.
\newblock In \emph{ICML}, pages 2790--2799, 2019.

\bibitem[Hu et~al.(2022)Hu, yelong shen, Wallis, Allen-Zhu, Li, Wang, Wang, and Chen]{hu2022lora}
Edward~J Hu, yelong shen, Phillip Wallis, Zeyuan Allen-Zhu, Yuanzhi Li, Shean Wang, Lu Wang, and Weizhu Chen.
\newblock Lo{RA}: Low-rank adaptation of large language models.
\newblock In \emph{ICLR}, 2022.

\bibitem[Hubara et~al.(2017)Hubara, Courbariaux, Soudry, El-Yaniv, and Bengio]{hubara2017quantized}
Itay Hubara, Matthieu Courbariaux, Daniel Soudry, Ran El-Yaniv, and Yoshua Bengio.
\newblock Quantized neural networks: Training neural networks with low precision weights and activations.
\newblock \emph{JMLR}, 18\penalty0 (1):\penalty0 6869--6898, 2017.

\bibitem[Izmailov et~al.(2018)Izmailov, Podoprikhin, Garipov, Vetrov, and Wilson]{izmailov2018averaging}
Pavel Izmailov, Dmitrii Podoprikhin, Timur Garipov, Dmitry Vetrov, and Andrew~Gordon Wilson.
\newblock Averaging weights leads to wider optima and better generalization.
\newblock In \emph{UAI}, 2018.

\bibitem[Jia et~al.(2022)Jia, Tang, Chen, Cardie, Belongie, Hariharan, and Lim]{jia2022vpt}
Menglin Jia, Luming Tang, Bor-Chun Chen, Claire Cardie, Serge Belongie, Bharath Hariharan, and Ser-Nam Lim.
\newblock Visual prompt tuning.
\newblock In \emph{ECCV}, 2022.

\bibitem[Khosla et~al.(2011)Khosla, Jayadevaprakash, Yao, and Fei-Fei]{Khosla_FGVC2011dogs}
Aditya Khosla, Nityananda Jayadevaprakash, Bangpeng Yao, and Li Fei-Fei.
\newblock Novel dataset for fine-grained image categorization.
\newblock In \emph{CVPRW}, 2011.

\bibitem[Kornblith et~al.(2019)Kornblith, Shlens, and Le]{kornblith2019better}
Simon Kornblith, Jonathon Shlens, and Quoc~V Le.
\newblock Do better imagenet models transfer better?
\newblock In \emph{CVPR}, pages 2661--2671, 2019.

\bibitem[Krizhevsky(2009)]{Krizhevsky09learningmultiple}
Alex Krizhevsky.
\newblock Learning multiple layers of features from tiny images.
\newblock Technical report, 2009.

\bibitem[Kundu et~al.(2023)Kundu, Wynter, Lee, and Bathen]{kundu2023tofa}
Achintya Kundu, Laura Wynter, Rhui~Dih Lee, and Luis~Angel Bathen.
\newblock Tofa: Transfer-once-for-all.
\newblock \emph{arXiv preprint arXiv:2303.15485}, 2023.

\bibitem[LeCun et~al.(1989)LeCun, Denker, and Solla]{lecun1989optimal}
Yann LeCun, John Denker, and Sara Solla.
\newblock Optimal brain damage.
\newblock \emph{NeurIPS}, 2, 1989.

\bibitem[Lee et~al.(2019)Lee, Ajanthan, and Torr]{lee2018snip}
Namhoon Lee, Thalaiyasingam Ajanthan, and Philip Torr.
\newblock Snip: Single-shot network pruning based on connection sensitivity.
\newblock In \emph{ICLR}, 2019.

\bibitem[Lenc and Vedaldi(2015)]{lenc2015understanding}
Karel Lenc and Andrea Vedaldi.
\newblock Understanding image representations by measuring their equivariance and equivalence.
\newblock In \emph{CVPR}, pages 991--999, 2015.

\bibitem[Lester et~al.(2021)Lester, Al-Rfou, and Constant]{lester2021power}
Brian Lester, Rami Al-Rfou, and Noah Constant.
\newblock The power of scale for parameter-efficient prompt tuning.
\newblock In \emph{EMNLP}, 2021.

\bibitem[Li and Liang(2021)]{li2021prefix}
Xiang~Lisa Li and Percy Liang.
\newblock Prefix-tuning: Optimizing continuous prompts for generation.
\newblock In \emph{ACL}, 2021.

\bibitem[Lian et~al.(2022)Lian, Zhou, Feng, and Wang]{lian2022scaling}
Dongze Lian, Daquan Zhou, Jiashi Feng, and Xinchao Wang.
\newblock Scaling \& shifting your features: A new baseline for efficient model tuning.
\newblock \emph{NeurIPS}, 2022.

\bibitem[Liu et~al.(2019)Liu, Peng, and Schwing]{liu2019knowledge}
Iou-Jen Liu, Jian Peng, and Alexander~G Schwing.
\newblock Knowledge flow: Improve upon your teachers.
\newblock In \emph{ICLR}, 2019.

\bibitem[Liu et~al.(2022{\natexlab{a}})Liu, Cai, and Zhuang]{liu2022focusformer}
Jing Liu, Jianfei Cai, and Bohan Zhuang.
\newblock Focusformer: Focusing on what we need via architecture sampler.
\newblock \emph{arXiv preprint arXiv:2208.10861}, 2022{\natexlab{a}}.

\bibitem[Liu et~al.(2023)Liu, Yuan, Fu, Jiang, Hayashi, and Neubig]{liu2023pretrain}
Pengfei Liu, Weizhe Yuan, Jinlan Fu, Zhengbao Jiang, Hiroaki Hayashi, and Graham Neubig.
\newblock Pre-train, prompt, and predict: A systematic survey of prompting methods in natural language processing.
\newblock \emph{ACM Comput. Surv.}, 2023.

\bibitem[Liu et~al.(2022{\natexlab{b}})Liu, Ma, Tian, He, and Kira]{liu2022polyhistor}
Yen-Cheng Liu, Chih-Yao Ma, Junjiao Tian, Zijian He, and Zsolt Kira.
\newblock Polyhistor: Parameter-efficient multi-task adaptation for dense vision tasks.
\newblock \emph{NeurIPS}, 2022{\natexlab{b}}.

\bibitem[Liu et~al.(2022{\natexlab{c}})Liu, Mao, Wu, Feichtenhofer, Darrell, and Xie]{liu2022convnet}
Zhuang Liu, Hanzi Mao, Chao-Yuan Wu, Christoph Feichtenhofer, Trevor Darrell, and Saining Xie.
\newblock A convnet for the 2020s.
\newblock In \emph{CVPR}, pages 11976--11986, 2022{\natexlab{c}}.

\bibitem[Loshchilov and Hutter(2017)]{loshchilov2017decoupled}
Ilya Loshchilov and Frank Hutter.
\newblock Decoupled weight decay regularization.
\newblock \emph{arXiv preprint arXiv:1711.05101}, 2017.

\bibitem[Lu et~al.(2021)Lu, Sreekumar, Goodman, Banzhaf, Deb, and Boddeti]{lu2021neural}
Zhichao Lu, Gautam Sreekumar, Erik Goodman, Wolfgang Banzhaf, Kalyanmoy Deb, and Vishnu~Naresh Boddeti.
\newblock Neural architecture transfer.
\newblock \emph{TPAMI}, 43\penalty0 (9):\penalty0 2971--2989, 2021.

\bibitem[Matena and Raffel(2022)]{matena2022merging}
Michael~S Matena and Colin~A Raffel.
\newblock Merging models with fisher-weighted averaging.
\newblock \emph{NeurIPS}, 35:\penalty0 17703--17716, 2022.

\bibitem[Nguyen et~al.(2020)Nguyen, Hassner, Seeger, and Archambeau]{nguyen2020leep}
Cuong Nguyen, Tal Hassner, Matthias Seeger, and Cedric Archambeau.
\newblock Leep: A new measure to evaluate transferability of learned representations.
\newblock In \emph{ICML}, pages 7294--7305. PMLR, 2020.

\bibitem[Nilsback and Zisserman(2008)]{nilsback2008automated}
Maria-Elena Nilsback and Andrew Zisserman.
\newblock Automated flower classification over a large number of classes.
\newblock In \emph{ICVGIP}, pages 722--729. IEEE, 2008.

\bibitem[Pan et~al.(2023)Pan, Cai, and Zhuang]{pan2023stitchable}
Zizheng Pan, Jianfei Cai, and Bohan Zhuang.
\newblock Stitchable neural networks.
\newblock In \emph{CVPR}, 2023.

\bibitem[Pham et~al.(2018)Pham, Guan, Zoph, Le, and Dean]{pham2018efficient}
Hieu Pham, Melody Guan, Barret Zoph, Quoc Le, and Jeff Dean.
\newblock Efficient neural architecture search via parameters sharing.
\newblock In \emph{ICML}, pages 4095--4104. PMLR, 2018.

\bibitem[Plested and Gedeon(2022)]{plested2022deep}
Jo Plested and Tom Gedeon.
\newblock Deep transfer learning for image classification: a survey.
\newblock \emph{arXiv preprint arXiv:2205.09904}, 2022.

\bibitem[Radosavovic et~al.(2020)Radosavovic, Kosaraju, Girshick, He, and Doll{\'a}r]{radosavovic2020designing}
Ilija Radosavovic, Raj~Prateek Kosaraju, Ross Girshick, Kaiming He, and Piotr Doll{\'a}r.
\newblock Designing network design spaces.
\newblock In \emph{CVPR}, pages 10428--10436, 2020.

\bibitem[Rombach et~al.(2022)Rombach, Blattmann, Lorenz, Esser, and Ommer]{rombach2022high}
Robin Rombach, Andreas Blattmann, Dominik Lorenz, Patrick Esser, and Bj{\"o}rn Ommer.
\newblock High-resolution image synthesis with latent diffusion models.
\newblock In \emph{CVPR}, pages 10684--10695, 2022.

\bibitem[Russakovsky et~al.(2015)Russakovsky, Deng, Su, Krause, Satheesh, Ma, Huang, Karpathy, Khosla, Bernstein, et~al.]{russakovsky2015imagenet}
Olga Russakovsky, Jia Deng, Hao Su, Jonathan Krause, Sanjeev Satheesh, Sean Ma, Zhiheng Huang, Andrej Karpathy, Aditya Khosla, Michael Bernstein, et~al.
\newblock Imagenet large scale visual recognition challenge.
\newblock \emph{IJCV}, 115:\penalty0 211--252, 2015.

\bibitem[Rusu et~al.(2016)Rusu, Rabinowitz, Desjardins, Soyer, Kirkpatrick, Kavukcuoglu, Pascanu, and Hadsell]{rusu2016progressive}
Andrei~A Rusu, Neil~C Rabinowitz, Guillaume Desjardins, Hubert Soyer, James Kirkpatrick, Koray Kavukcuoglu, Razvan Pascanu, and Raia Hadsell.
\newblock Progressive neural networks.
\newblock \emph{arXiv preprint arXiv:1606.04671}, 2016.

\bibitem[Shu et~al.(2021)Shu, Kou, Cao, Wang, and Long]{shu2021zoo}
Yang Shu, Zhi Kou, Zhangjie Cao, Jianmin Wang, and Mingsheng Long.
\newblock Zoo-tuning: Adaptive transfer from a zoo of models.
\newblock In \emph{ICML}, pages 9626--9637. PMLR, 2021.

\bibitem[Shu et~al.(2022)Shu, Cao, Zhang, Wang, and Long]{shu2022hub}
Yang Shu, Zhangjie Cao, Ziyang Zhang, Jianmin Wang, and Mingsheng Long.
\newblock Hub-pathway: Transfer learning from a hub of pre-trained models.
\newblock \emph{NeurIPS}, 2022.

\bibitem[Steiner et~al.(2022)Steiner, Kolesnikov, Zhai, Wightman, Uszkoreit, and Beyer]{steiner2021train}
Andreas Steiner, Alexander Kolesnikov, Xiaohua Zhai, Ross Wightman, Jakob Uszkoreit, and Lucas Beyer.
\newblock How to train your vit? data, augmentation, and regularization in vision transformers.
\newblock \emph{TMLR}, 2022.

\bibitem[Sung et~al.(2022)Sung, Cho, and Bansal]{sung2022lst}
Yi-Lin Sung, Jaemin Cho, and Mohit Bansal.
\newblock Lst: Ladder side-tuning for parameter and memory efficient transfer learning.
\newblock \emph{NeurIPS}, 2022.

\bibitem[Taori et~al.(2023)Taori, Gulrajani, Zhang, Dubois, Li, Guestrin, Liang, and Hashimoto]{alpaca}
Rohan Taori, Ishaan Gulrajani, Tianyi Zhang, Yann Dubois, Xuechen Li, Carlos Guestrin, Percy Liang, and Tatsunori~B. Hashimoto.
\newblock Stanford alpaca: An instruction-following llama model.
\newblock \url{https://github.com/tatsu-lab/stanford_alpaca}, 2023.

\bibitem[Touvron et~al.(2021)Touvron, Cord, Douze, Massa, Sablayrolles, and J{\'e}gou]{touvron2021training}
Hugo Touvron, Matthieu Cord, Matthijs Douze, Francisco Massa, Alexandre Sablayrolles, and Herv{\'e} J{\'e}gou.
\newblock Training data-efficient image transformers \& distillation through attention.
\newblock In \emph{ICML}, pages 10347--10357. PMLR, 2021.

\bibitem[Touvron et~al.(2023{\natexlab{a}})Touvron, Lavril, Izacard, Martinet, Lachaux, Lacroix, Rozi{\`e}re, Goyal, Hambro, Azhar, et~al.]{touvron2023llama}
Hugo Touvron, Thibaut Lavril, Gautier Izacard, Xavier Martinet, Marie-Anne Lachaux, Timoth{\'e}e Lacroix, Baptiste Rozi{\`e}re, Naman Goyal, Eric Hambro, Faisal Azhar, et~al.
\newblock Llama: Open and efficient foundation language models.
\newblock \emph{arXiv preprint arXiv:2302.13971}, 2023{\natexlab{a}}.

\bibitem[Touvron et~al.(2023{\natexlab{b}})Touvron, Martin, Stone, Albert, Almahairi, Babaei, Bashlykov, Batra, Bhargava, Bhosale, et~al.]{touvron2023llama2}
Hugo Touvron, Louis Martin, Kevin Stone, Peter Albert, Amjad Almahairi, Yasmine Babaei, Nikolay Bashlykov, Soumya Batra, Prajjwal Bhargava, Shruti Bhosale, et~al.
\newblock Llama 2: Open foundation and fine-tuned chat models.
\newblock \emph{arXiv preprint arXiv:2307.09288}, 2023{\natexlab{b}}.

\bibitem[Tran et~al.(2019)Tran, Nguyen, and Hassner]{tran2019transferability}
Anh~T Tran, Cuong~V Nguyen, and Tal Hassner.
\newblock Transferability and hardness of supervised classification tasks.
\newblock In \emph{ICCV}, pages 1395--1405, 2019.

\bibitem[Van~Horn et~al.(2015)Van~Horn, Branson, Farrell, Haber, Barry, Ipeirotis, Perona, and Belongie]{van2015building}
Grant Van~Horn, Steve Branson, Ryan Farrell, Scott Haber, Jessie Barry, Panos Ipeirotis, Pietro Perona, and Serge Belongie.
\newblock Building a bird recognition app and large scale dataset with citizen scientists: The fine print in fine-grained dataset collection.
\newblock In \emph{CVPR}, pages 595--604, 2015.

\bibitem[Wah et~al.(2011)Wah, Branson, Welinder, Perona, and Belongie]{wah2011caltech}
Catherine Wah, Steve Branson, Peter Welinder, Pietro Perona, and Serge Belongie.
\newblock The caltech-ucsd birds-200-2011 dataset.
\newblock \emph{Tech. Rep. CNS-TR-2011-001, California Institute of Technology}, 2011.

\bibitem[Wang et~al.(2020)Wang, Zhang, and Grosse]{wang2020picking}
Chaoqi Wang, Guodong Zhang, and Roger Grosse.
\newblock Picking winning tickets before training by preserving gradient flow.
\newblock In \emph{ICLR}, 2020.

\bibitem[Wang et~al.(2021)Wang, Li, Gong, and Chandra]{wang2021attentivenas}
Dilin Wang, Meng Li, Chengyue Gong, and Vikas Chandra.
\newblock Attentivenas: Improving neural architecture search via attentive sampling.
\newblock In \emph{CVPR}, pages 6418--6427, 2021.

\bibitem[Wang et~al.(2022)Wang, Kordi, Mishra, Liu, Smith, Khashabi, and Hajishirzi]{wang2022self}
Yizhong Wang, Yeganeh Kordi, Swaroop Mishra, Alisa Liu, Noah~A Smith, Daniel Khashabi, and Hannaneh Hajishirzi.
\newblock Self-instruct: Aligning language model with self generated instructions.
\newblock \emph{arXiv preprint arXiv:2212.10560}, 2022.

\bibitem[Wolf et~al.(2020)Wolf, Debut, Sanh, Chaumond, Delangue, Moi, Cistac, Rault, Louf, Funtowicz, et~al.]{wolf2020transformers}
Thomas Wolf, Lysandre Debut, Victor Sanh, Julien Chaumond, Clement Delangue, Anthony Moi, Pierric Cistac, Tim Rault, R{\'e}mi Louf, Morgan Funtowicz, et~al.
\newblock Transformers: State-of-the-art natural language processing.
\newblock In \emph{EMNLP}, pages 38--45, 2020.

\bibitem[Wortsman et~al.(2022{\natexlab{a}})Wortsman, Ilharco, Gadre, Roelofs, Gontijo-Lopes, Morcos, Namkoong, Farhadi, Carmon, Kornblith, et~al.]{wortsman2022model}
Mitchell Wortsman, Gabriel Ilharco, Samir~Ya Gadre, Rebecca Roelofs, Raphael Gontijo-Lopes, Ari~S Morcos, Hongseok Namkoong, Ali Farhadi, Yair Carmon, Simon Kornblith, et~al.
\newblock Model soups: averaging weights of multiple fine-tuned models improves accuracy without increasing inference time.
\newblock In \emph{ICML}, pages 23965--23998. PMLR, 2022{\natexlab{a}}.

\bibitem[Wortsman et~al.(2022{\natexlab{b}})Wortsman, Ilharco, Kim, Li, Kornblith, Roelofs, Lopes, Hajishirzi, Farhadi, Namkoong, et~al.]{wortsman2022robust}
Mitchell Wortsman, Gabriel Ilharco, Jong~Wook Kim, Mike Li, Simon Kornblith, Rebecca Roelofs, Raphael~Gontijo Lopes, Hannaneh Hajishirzi, Ali Farhadi, Hongseok Namkoong, et~al.
\newblock Robust fine-tuning of zero-shot models.
\newblock In \emph{CVPR}, pages 7959--7971, 2022{\natexlab{b}}.

\bibitem[Xu et~al.(2021{\natexlab{a}})Xu, Kang, Zhang, Xie, Liang, and Li]{xu2021nasoa}
Hang Xu, Ning Kang, Gengwei Zhang, Chuanlong Xie, Xiaodan Liang, and Zhenguo Li.
\newblock Nasoa: Towards faster task-oriented online fine-tuning with a zoo of models.
\newblock In \emph{ICCV}, pages 5097--5106, 2021{\natexlab{a}}.

\bibitem[Xu et~al.(2021{\natexlab{b}})Xu, Luo, Zhang, Tan, Chang, Huang, and Huang]{xu2021raise}
Runxin Xu, Fuli Luo, Zhiyuan Zhang, Chuanqi Tan, Baobao Chang, Songfang Huang, and Fei Huang.
\newblock Raise a child in large language model: Towards effective and generalizable fine-tuning.
\newblock In \emph{EMNLP}, 2021{\natexlab{b}}.

\bibitem[Yang et~al.(2022)Yang, Zhou, Liu, Ye, and Wang]{yang2022deep}
Xingyi Yang, Daquan Zhou, Songhua Liu, Jingwen Ye, and Xinchao Wang.
\newblock Deep model reassembly.
\newblock \emph{NeurIPS}, 35:\penalty0 25739--25753, 2022.

\bibitem[Yosinski et~al.(2014)Yosinski, Clune, Bengio, and Lipson]{yosinski2014transferable}
Jason Yosinski, Jeff Clune, Yoshua Bengio, and Hod Lipson.
\newblock How transferable are features in deep neural networks?
\newblock \emph{NeurIPS}, 27, 2014.

\bibitem[You et~al.(2021)You, Liu, Wang, and Long]{you2021logme}
Kaichao You, Yong Liu, Jianmin Wang, and Mingsheng Long.
\newblock Logme: Practical assessment of pre-trained models for transfer learning.
\newblock In \emph{ICML}, pages 12133--12143. PMLR, 2021.

\bibitem[Yu and Huang(2019)]{yu2019universally}
Jiahui Yu and Thomas~S Huang.
\newblock Universally slimmable networks and improved training techniques.
\newblock In \emph{ICCV}, pages 1803--1811, 2019.

\bibitem[Zaken et~al.(2022)Zaken, Goldberg, and Ravfogel]{zaken2022bitfit}
Elad~Ben Zaken, Yoav Goldberg, and Shauli Ravfogel.
\newblock Bitfit: Simple parameter-efficient fine-tuning for transformer-based masked language-models.
\newblock In \emph{ACL}, pages 1--9, 2022.

\bibitem[Zhai et~al.(2019)Zhai, Puigcerver, Kolesnikov, Ruyssen, Riquelme, Lucic, Djolonga, Pinto, Neumann, Dosovitskiy, et~al.]{zhai2019vtab}
Xiaohua Zhai, Joan Puigcerver, Alexander Kolesnikov, Pierre Ruyssen, Carlos Riquelme, Mario Lucic, Josip Djolonga, Andre~Susano Pinto, Maxim Neumann, Alexey Dosovitskiy, et~al.
\newblock A large-scale study of representation learning with the visual task adaptation benchmark.
\newblock \emph{arXiv preprint arXiv:1910.04867}, 2019.

\bibitem[Zhang and Agrawala(2023)]{zhang2023adding}
Lvmin Zhang and Maneesh Agrawala.
\newblock Adding conditional control to text-to-image diffusion models.
\newblock In \emph{ICCV}, 2023.

\bibitem[Zhang et~al.(2023)Zhang, Han, Zhou, Hu, Yan, Lu, Li, Gao, and Qiao]{zhang2023llama}
Renrui Zhang, Jiaming Han, Aojun Zhou, Xiangfei Hu, Shilin Yan, Pan Lu, Hongsheng Li, Peng Gao, and Yu Qiao.
\newblock Llama-adapter: Efficient fine-tuning of language models with zero-init attention.
\newblock \emph{arXiv preprint arXiv:2303.16199}, 2023.

\bibitem[Zhang et~al.(2022)Zhang, Zhou, and Liu]{zhang2022neural}
Yuanhan Zhang, Kaiyang Zhou, and Ziwei Liu.
\newblock Neural prompt search.
\newblock \emph{arXiv preprint arXiv:2206.04673}, 2022.

\bibitem[Zhou et~al.(2022)Zhou, Yang, Loy, and Liu]{zhou2022conditional}
Kaiyang Zhou, Jingkang Yang, Chen~Change Loy, and Ziwei Liu.
\newblock Conditional prompt learning for vision-language models.
\newblock In \emph{CVPR}, pages 16816--16825, 2022.

\end{thebibliography}
}
\appendix

\clearpage
\begin{center}
    \Large{\textbf{Appendix}}
\end{center}

We organize the appendix as follows. 

\begin{figure*}[htb]
\begin{center}
\includegraphics[width=\linewidth]{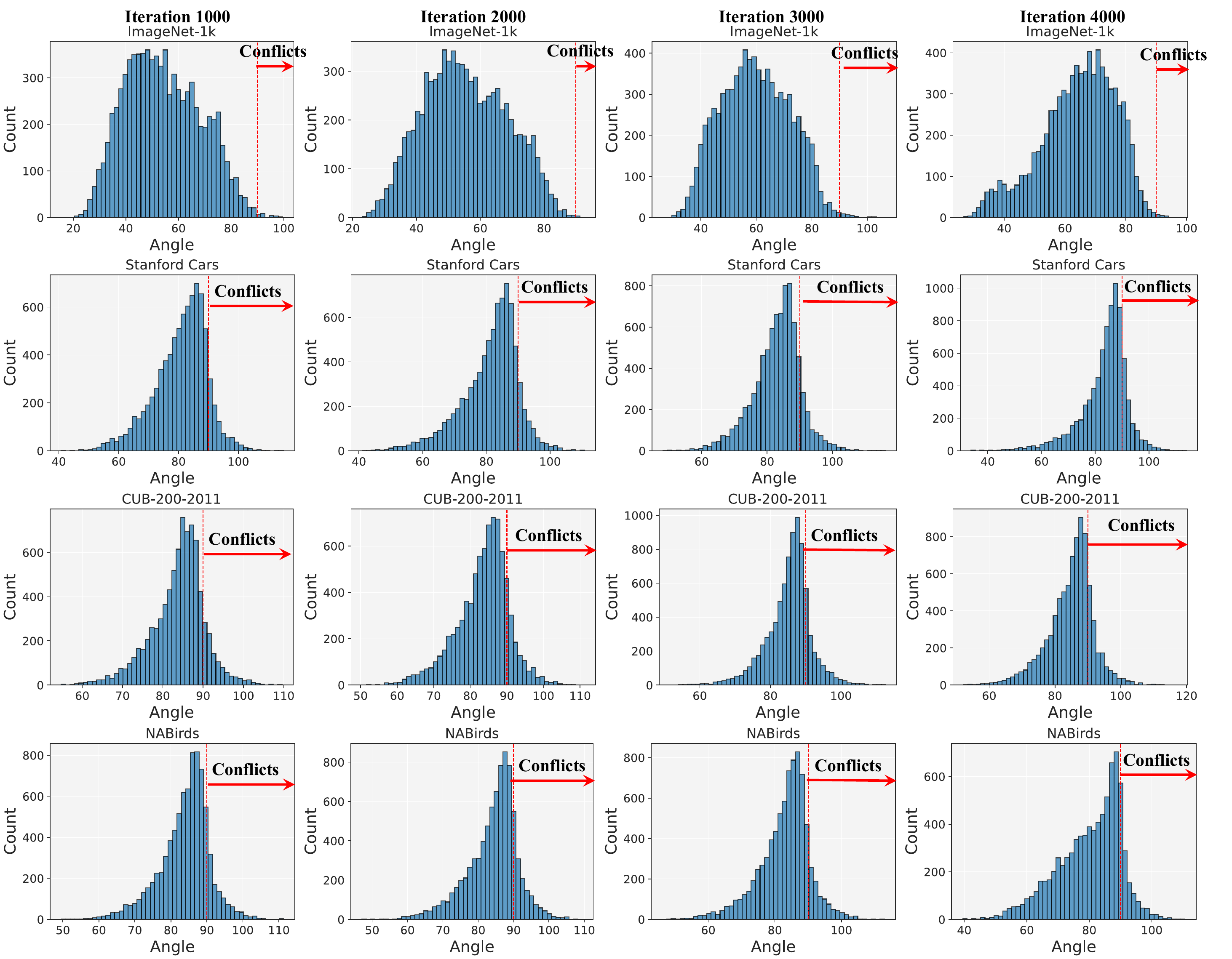}
\end{center}
\caption{More visualizations for distributions of pair-wise gradient angles among stitches when updating shared weights. We show visualizations for ImageNet-1k~\cite{russakovsky2015imagenet}, Stanford Cars~\cite{gebru2017cars}, CUB-200-2011~\cite{wah2011caltech}, and NABirds~\cite{van2015building} datasets. Angle 90$^\circ$ is highlighted with a dashed red line. For simplicity, we combine the gradients for query, key, and value projections and show their angles 
for a total of 32 stitches when stitching ViT-Ti and ViT-S anchors. Similar to Figure~\ref{fig:bias_motivation} of the main paper, the gradient angles are larger in the downstream datasets than in the source domain ImageNet-1k~\cite{russakovsky2015imagenet}.}
\label{fig:bias_motivation_more}
\end{figure*}

\begin{itemize}[itemsep=5pt]
    \item In Section~\ref{subsec:more_hyper}, we introduce more details about data augmentation and other hyper-parameters.
    \item In Section~\ref{subsec:few_shot}, we show few-shot learning performance comparisons with SN-Net on Stanford Cars~\cite{gebru2017cars}.
    \item In Section~\ref{subsec:conv}, we adapt our ESTA to convnets and show the performance comparisons with SN-Net and anchors individually fine-tuned with LoRA~\cite{hu2022lora}.
    \item In Section~\ref{subsec:dist}, we investigate the effectiveness of our distillation strategy.
    \item In Section~\ref{subsec:eta}, we study the effect of the momentum coefficient $\eta$.
    \item In Section~\ref{subsec:adapt_and_stitch}, we analyze why our strategy of simultaneously stitching and adapting anchors leads to better performance.
    \item In Section~\ref{subsec:pattern}, we investigate whether the deployed task-specific stitches are the same across three sampled tasks: Oxford Flowers~\cite{nilsback2008automated}, Stanford Cars~\cite{gebru2017cars}, and CUB-200-2011~\cite{wah2011caltech}.
    \item \hy{In Section~\ref{subsec:compare_lora}, we conduct more experiments comparing our proposed PST with naive LoRA.}
    \item In Figure~\ref{fig:bias_motivation_more}, we show more visualizations for distributions of pair-wise gradient angles among stitches when updating shared weights similar to Figure~\ref{fig:bias_motivation} of the main paper.
    \item In Table~\ref{tab:quantitative_fig34}, we show quantitative results from Figures~\ref{fig:main_results} and~\ref{fig:main_vtab} of the main paper.
    \item In Section~\ref{subsec:instruction_qualitative}, we show more qualitative results on the instruction-following task.
\end{itemize}

\section{Data Augmentation and Other Hyper-parameters} \label{subsec:more_hyper}
For our ESTA and the baseline methods, we follow~\cite{jia2022vpt} for a standard data augmentation pipeline with resize, random crop to 224$\times$224 and random horizontal flip for the FGVC and CIFAR-100 datasets, and only resize to 224$\times$224 for VTAB-1k datasets. Following~\cite{zhang2022neural}, we use AdamW optimizer~\cite{loshchilov2017decoupled} with a cosine scheduler and a learning rate warm-up period of 10 epochs.
We set weight decay as $1\times 10^{-4}$ for all visual recognition datasets. We choose the base learning rate with a grid search on the validation set over \{$1\times 10^{-4}$, $2\times 10^{-4}$, $1\times 10^{-3}$, $2\times 10^{-3}$\} following the split of~\cite{jia2022vpt}. We set the learning rate as $base\_lr\times b/256$, where $b$ is the batch size. All experiments are conducted on NVIDIA A100 GPUs. Due to the constraints of computational resources, we use 8-bit matrix multiplication~\cite{dettmers2022llm} for the frozen weights in the instruction-following task during fine-tuning and generation.

\begin{figure*}[t]
\begin{center}
\includegraphics[width=\linewidth]{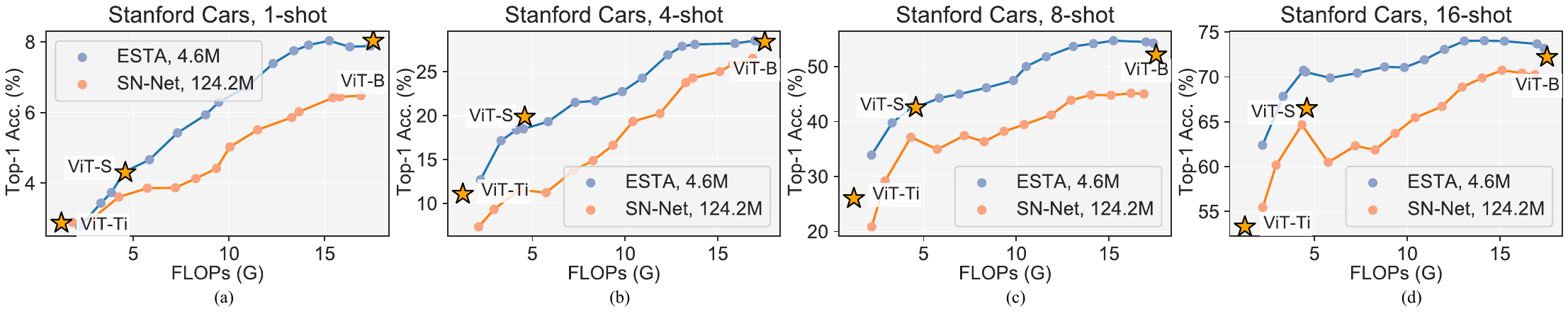}
\end{center}
\caption{Few-shot learning performance comparisons with SN-Net~\cite{pan2023stitchable} for adapting ViT-Ti/S/B pre-trained on ImageNet-22k~\cite{deng2009imagenet} to Stanford Cars~\cite{gebru2017cars}. We denote individually fine-tuned anchors as yellow stars and also show the number of trainable parameters.}
\label{fig:few_shot}
\end{figure*}

\section{Results on Few-shot Learning}\label{subsec:few_shot}
Following~\cite{zhou2022conditional}, we conduct experiments with limited samples (1, 4, 8, 16 shots) on Stanford Cars task~\cite{gebru2017cars}. The results are visualized in Figure~\ref{fig:few_shot}. We have similar observations as those in Section~\ref{subsec:visual_main} of the main paper that our ESTA 1) exhibits smooth FLOPs-accuracy curves; 2) generally outperforms SN-Net by large margins; and 3) achieves comparable or even better performance than individually fine-tuned anchors, which we conjecture due to model stitching’s strong weight-sharing regularization. This suggests that our ESTA works well in data-efficient adaptation.

\begin{figure*}[t]
\begin{center}
\includegraphics[width=0.75\linewidth]{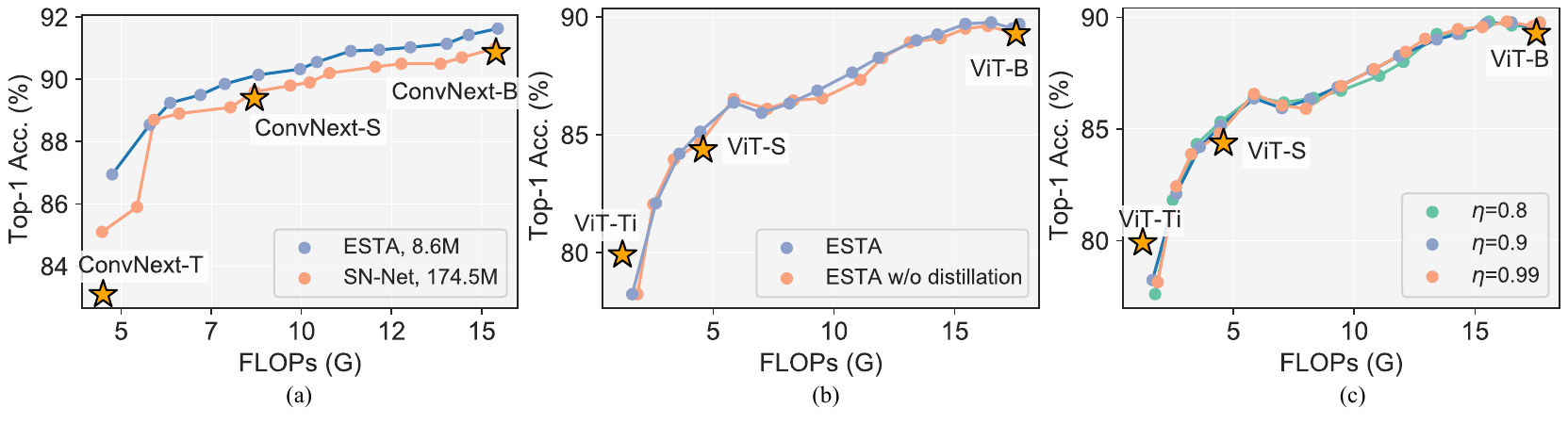}
\end{center}
\caption{(a) Comparisons with SN-Net~\cite{pan2023stitchable} for adapting
ConvNext-T/S/B~\cite{liu2022convnet} pre-trained on ImageNet-22k~\cite{deng2009imagenet}. We denote individually fine-tuned anchors as
yellow stars and also show the number of trainable parameters. (b) Effect of inplace distillation. (c) Effect of the momentum coefficient $\eta$ for updating the importance scores. We show averaged results on five FGVC datasets for (a), (b), and (c).}
\label{fig:supp_effect_conv}
\end{figure*}

\section{Apply ESTA to Convnets}\label{subsec:conv}
We employ our ESTA to adapt the popular convnet architecture ConvNext-T/S/B~\cite{liu2022convnet} pre-trained on ImageNet-22k. The averaged results on five FGVC datasets are visualized in Figure~\ref{fig:supp_effect_conv} (a). We have similar observations as those in Figures~\ref{fig:main_results} and~\ref{fig:main_vtab} of the main paper that our ESTA obtains stitches with a smooth FLOPs-accuracy curve and outperforms SN-Net by large margins. This indicates that our ESTA is not restricted to transformer-based architectures.

\section{Effect of Inplace Distillation}\label{subsec:dist}
We investigate the effect of inplace distillation that is introduced in Section~\ref{subsec:one_stage} of the main paper. The averaged results on five FGVC tasks are visualized in Figure~\ref{fig:supp_effect_conv} (b). We observe that inplace distillation contributes to the superiority of our proposed ESTA framework for obtaining stitches of good performance. However, its contribution is not as significant as the key designs of ESTA, \ie, our parameter-efficient stitch fine-tuning method, a one-stage deployment pipeline with a task-specific stitch sampling strategy as shown in Figure~\ref{fig:abl_effect} of the main paper.

\section{Effect of $\eta$}\label{subsec:eta}
As introduced in Section~\ref{subsec:one_stage} of the main paper, $\eta$ is the momentum coefficient for updating the importance scores of the stitches. We investigate the effect of $\eta$ and visualize the averaged results on five FGVC datasets in Figure~\ref{fig:supp_effect_conv} (c). We empirically find that setting $\eta$ to 0.9 and 0.99 achieves slightly better results than setting $\eta$ to 0.8. We conjecture that a small value of $\eta$ makes the sampling process unstable during an early training stage after the short warm-up period. Therefore, we set $\eta$ to 0.9 as the default setting.

\begin{figure*}[t]
\begin{center}
\includegraphics[width=\linewidth]{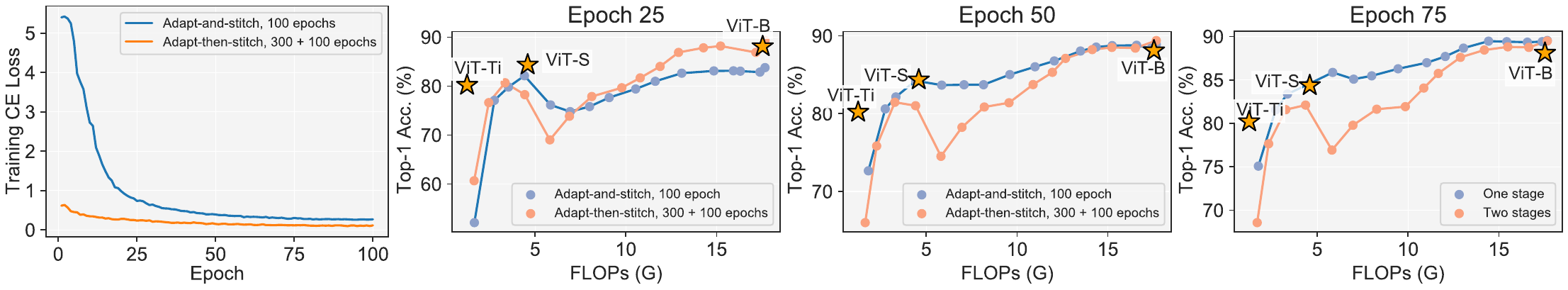}
\end{center}
\caption{Fine-tuning cross-entropy loss and testing accuracy with respect to the fine-tuning epochs averaged over five FGVC datasets. We show the testing accuracy for 25, 50, and 75 epochs. Our strategy ``Adapt-and-stitch'' takes 100 epochs for fine-tuning. In contrast, ``Adapt-then-stitch'', as a straightforward approach to apply SN-Net, first individually adapts each anchor for a total of $300$ epochs, and then fine-tunes SN-Net for another 100 epochs.}
\label{fig:training_vs_testing}
\end{figure*}

\section{Why our ``Adapt-and-stitch'' leads to better performance?}\label{subsec:adapt_and_stitch}
As shown in Section~\ref{subsec:abl} and Figure~\ref{fig:abl_effect} (d) of the main paper, our strategy to simultaneously adapt and stitch anchors achieves significant performance gain from SN-Net's straightforward strategy to first adapt and then stitch anchors. We speculate that anchors are likely to overfit during fine-tuning in SN-Net’s anchor adaptation stage and lose their generalization capability~\cite{wortsman2022robust,andreassen2021evolution}. Accordingly, stitching their weights in SN-Net may not generalize well to unseen samples. To verify our speculation, we compare the training cross-entropy loss and testing accuracy with respect to the training epoch. In Figure~\ref{fig:training_vs_testing}, we observe that SN-Net's straightforward ``Adapt-then-stitch'' strategy exhibits lower cross-entropy loss throughout training, but the testing accuracy barely improves after epoch 50. This observation aligns with our speculation and suggests the improved generalization capability of our ``Adapt-and-stitch'' pipeline.

\begin{figure*}[t]
\begin{center}
\includegraphics[width=\linewidth]{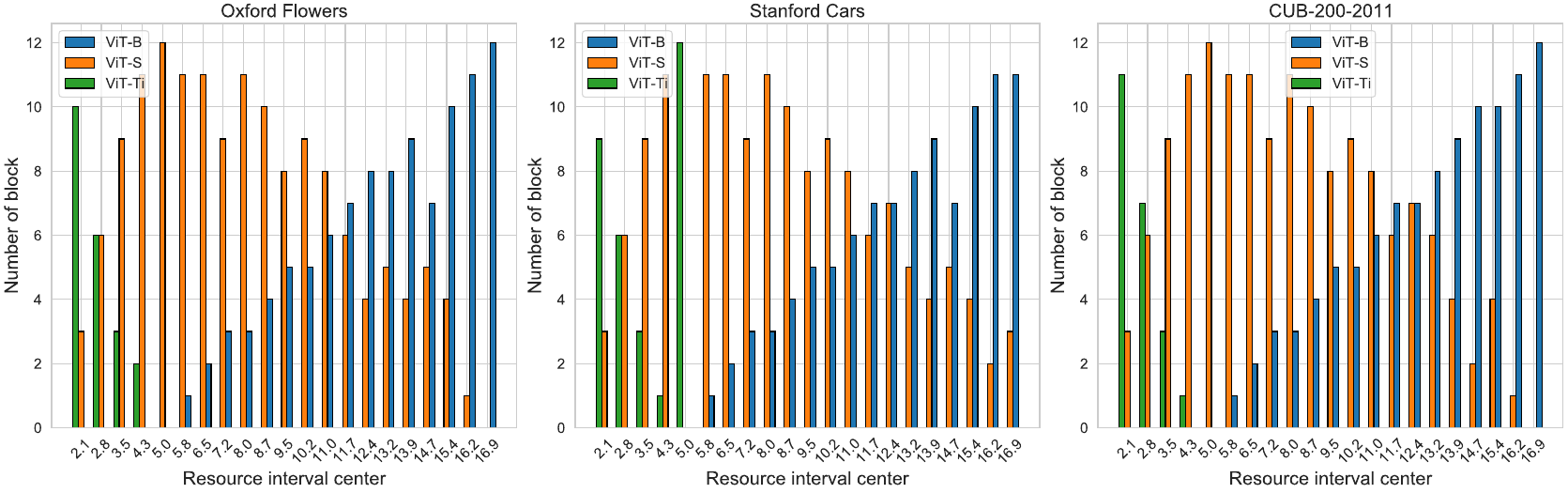}
\end{center}
\caption{The patterns of the deployed stitches selected by our task-specific sampling strategy on Oxford Flowers~\cite{nilsback2008automated}, Stanford Cars~\cite{gebru2017cars}, and CUB-200-2011~\cite{wah2011caltech} datasets. We visualize the number of blocks from anchors to form these stitches.}
\label{fig:pattern}
\end{figure*}

\begin{figure}[ht]
\begin{center}
\vspace{-1.em}
\includegraphics[width=\linewidth]{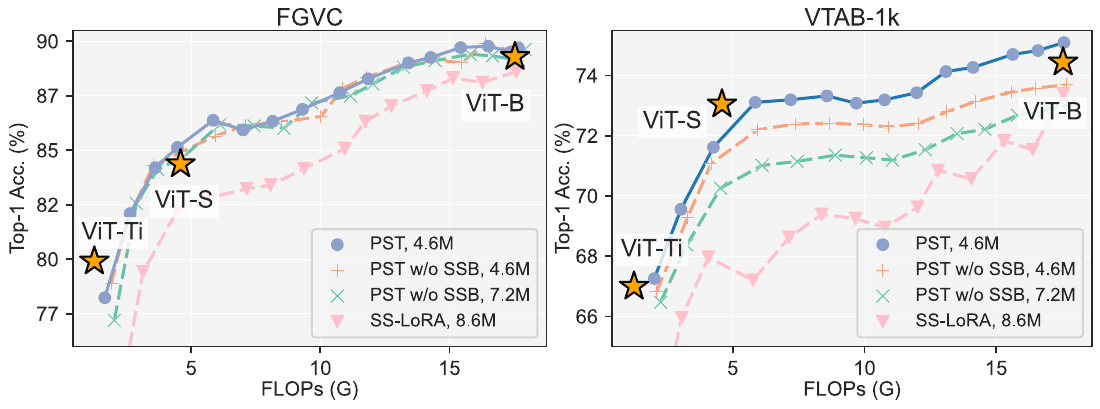}
\end{center}\vspace{-2.em}
\caption{\hy{Comparisons with naive LoRA and stitch-specific LoRA (SS-LoRA). We show averaged results on FGVC (left) and VTAB-1k (right). ``SSB'' represents stitch-specific bias terms}.}
\label{fig:abl_lora}
\vspace{-1.em}
\end{figure}

\begin{figure*}[ht]
\begin{center}
\includegraphics[width=\linewidth]{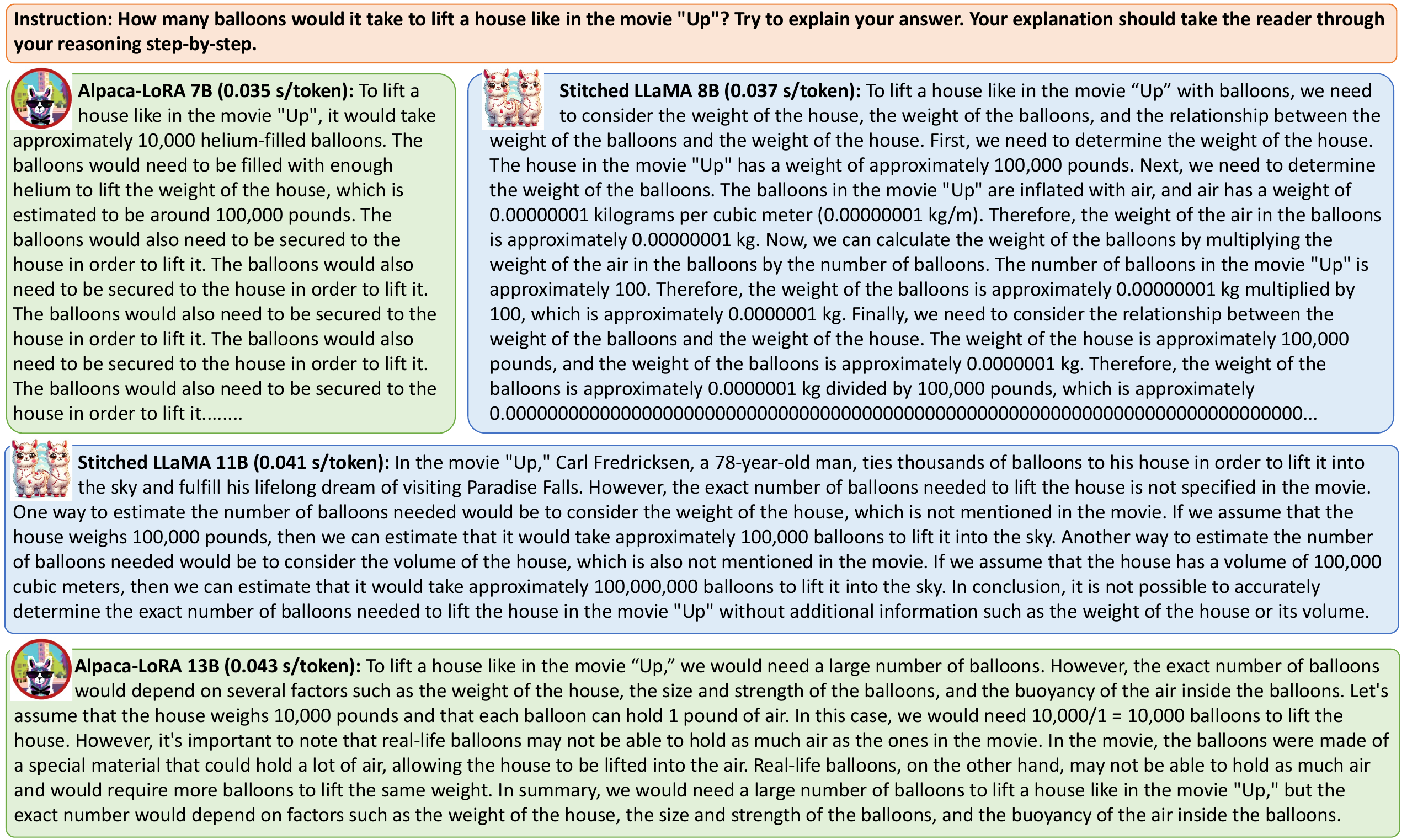}
\end{center}
\caption{Instruction-following comparison between Stitched LLaMA obtained by our ESTA and the individually fine-tuned anchors (Alpaca-LoRA 7B and 13B)~\cite{hu2022lora}.}
\label{fig:llama_vis_more1}
\end{figure*}

\begin{figure*}[ht]
\begin{center}
\includegraphics[width=\linewidth]{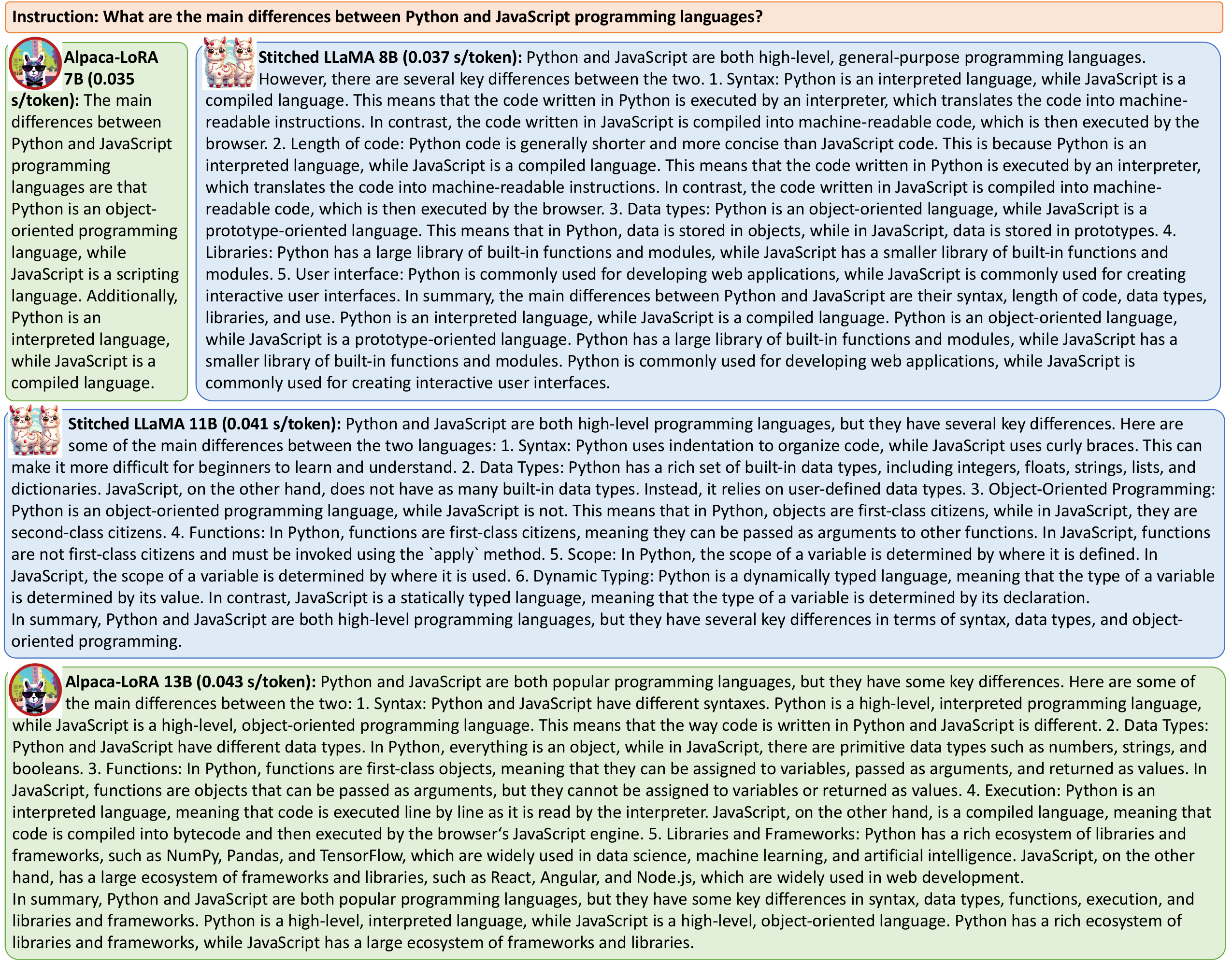}
\end{center}
\caption{Instruction-following comparison between Stitched LLaMA obtained by our ESTA and the individually fine-tuned anchors (Alpaca-LoRA 7B and 13B)~\cite{hu2022lora}.}
\label{fig:llama_vis_more2}
\end{figure*}

\section{Are important stitches the same across different tasks?}\label{subsec:pattern}
We investigate whether the important stitches are the same across tasks on Oxford Flowers~\cite{nilsback2008automated}, Stanford Cars~\cite{gebru2017cars}, and CUB-200-2011~\cite{wah2011caltech} datasets. The patterns are visualized in Figure~\ref{fig:pattern}. We observe that the selected important stitches are different on these visual recognition tasks, which verifies our motivation for deploying task-specific stitches and assigning higher sampling probabilities to them.

\section{\hy{More Comparisons with Naive LoRA}}\label{subsec:compare_lora}
\hy{We have shown that our PST excels naive LoRA (PST w/o stitch-specific bias) by a non-negligible margin in Figure~\ref{fig:abl_effect}~(b) of the main paper. In Figure~\ref{fig:abl_lora}, we further experiment on both FGVC and VTAB-1k benchmarks and compare PST with: 1) naive LoRA with different trainable parameters (different ranks); and 2) stitch-specific LoRA where we optimize an independent set of LoRA modules for each stitch. Across both benchmarks, our PST exceeds the competitors under comparable or lower trainable parameters, suggesting its superiority.}

\section{More Qualitative Evaluation on Instruction-following Task}\label{subsec:instruction_qualitative}
We show more qualitative comparisons between our Stitched LLaMA and the individually fine-tuned Alpaca-LoRA 7B and 13B in Figures~\ref{fig:llama_vis_more1} and~\ref{fig:llama_vis_more2}. We empirically find that the quality of the responses for our Stitched LLaMA improves with more parameters, which aligns with our findings in Sections~\ref{subsec:visual_main} and~\ref{subsec:llm} of the main paper. For instance, in Figure~\ref{fig:llama_vis_more1}, both Alpaca-LoRA 7B and Stitched LLaMA 8B tend to generate repetitive responses; however, Stitched LLaMA with higher numbers of parameters produce more coherent and informative responses. In addition, as shown in Figure~\ref{fig:llama_vis_more2}, as the total parameters of the stitches increase, the response content becomes richer and more detailed.

\begin{table*}[t]
\centering
\renewcommand\arraystretch{1.2}
\caption{Quantitative results from Figures~\ref{fig:main_results} and~\ref{fig:main_vtab} of the main paper.}
\vspace{-1em}
\label{tab:quantitative_fig34}
\resizebox{\linewidth}{!}{
\begin{tabular}{l|ccccccccccccccc} 
& \multicolumn{15}{c}{Stanford Cars~\cite{gebru2017cars}} \\\shline
FLOPs (G) & 2.48  &   3.31  &   4.41  &   5.85  &   6.93  &   8.39  &   9.47  &   10.54  &   11.62  &   13.77  &   14.16  &   15.54  &   16.61  &   17.39  &   17.67 \\
Accuracy (\%) & 82.00  &   84.37  &   85.25  &   84.87  &   85.40  &   86.11  &   86.22  &   86.83  &   86.99  &   86.15  &   87.58  &   87.68  &   87.90  &   87.28  &   87.20 \\\shline
& \multicolumn{15}{c}{CUB-200-2011~\cite{wah2011caltech}} \\\shline
FLOPs (G) & 2.86  &   3.97  &   4.52  &   5.85  &   6.93  &   8.00  &   9.08  &   10.54  &   11.62  &   13.77  &   14.46  &   15.93  &   17.00  &   17.39  &   17.67 \\
Accuracy (\%) & 81.29  &   82.86  &   83.62  &   85.54  &   84.36  &   85.43  &   86.05  &   86.71  &   86.88  &   88.06  &   87.76  &   88.44  &   88.16  &   87.88  &   88.47 \\\shline
& \multicolumn{15}{c}{Stanford Dogs~\cite{Khosla_FGVC2011dogs}} \\\shline
FLOPs (G) & 2.20  &   3.53  &   4.36  &   5.85  &   6.93  &   8.39  &   9.08  &   10.93  &   12.01  &   13.39  &   14.16  &   15.24  &   16.31  &   17.39  &   17.67 \\
Accuracy (\%) & 76.17  &   78.72  &   79.98  &   81.83  &   81.18  &   82.19  &   82.95  &   83.61  &   85.02  &   87.65  &   87.73  &   88.51  &   88.64  &   88.47  &   89.13 \\\shline
& \multicolumn{15}{c}{NABirds~\cite{van2015building}} \\\shline
FLOPs (G)  &   3.03  &   3.80  &   4.63  &   5.85  &   6.93  &   8.00  &   9.85  &   10.54  &   12.01  &   13.08  &   14.46  &   15.24  &   16.31  &   17.39  &   17.67 \\
Accuracy (\%)  &   77.23  &   78.28  &   79.15  &   81.71  &   80.79  &   81.85  &   83.22  &   83.65  &   84.09  &   84.44  &   84.20  &   85.12  &   85.19  &   85.14  &   85.06 \\\shline
& \multicolumn{15}{c}{VTAB~\cite{zhai2019vtab}} \\\shline
FLOPs (G)  &   2.02  &   3.03  &   4.26  &  4.53 &  5.85  &   7.20  &   8.57  &   9.69  &   10.76  &   11.98  &   13.10  &   14.12  &   15.63  &   16.59  &   17.57  \\
Accuracy (\%)  &   67.25  &   69.56  &   71.61 & 71.24  &   73.11  &   73.19  &   73.32  &   73.08  &   73.19  &   73.43  &   74.13  &   74.27  &   74.7  &   74.83  &   75.10  \\\shline
& \multicolumn{15}{c}{CIFAR-100~\cite{Krizhevsky09learningmultiple}} \\\shline
FLOPs (G)  &  2.35  &   3.08  &   4.37 & 4.65 &   5.83  &   8.76  &   9.84  & 10.92 &  12.00  &   12.7 &  13.08  &   14.16  &   15.24  &   15.94  &   17.67  \\
Accuracy (\%)  & 86.60  &   88.65  &   90.92 & 90.91  &   90.72  &   91.18  &   90.93 & 91.03  &   91.56  &   91.33  & 92.15 &   92.12  &   92.59  &   92.55  &   93.08  \\\shline
\end{tabular}
}
\end{table*}



\end{document}